\documentclass[10pt]{article}

\usepackage{scai}

\usepackage[authoryear, sort&compress, round]{natbib}
\usepackage{xspace}
\usepackage{adjustbox}
\usepackage{array}
\usepackage{float}
\usepackage{dblfloatfix}
\usepackage{setspace}
\usepackage{footmisc}
\usepackage{booktabs}
\usepackage{multirow}
\usepackage{longtable}
\usepackage{tabularx}
\usepackage{xltabular}
\usepackage{threeparttable}
\usepackage{siunitx}
\usepackage{mathtools}
\usepackage{bm}
\usepackage{dsfont}
\usepackage{subcaption}
\usepackage{tikz}
\usepackage{pgfplots}
\pgfplotsset{compat=1.18}
\usepackage{algorithm}
\usepackage{algpseudocode}
\usepackage{listings}
\usepackage{cleveref}
\usepackage{todonotes}
\usepackage{fontawesome5}
\usepackage{CJKutf8}
\usepackage{lipsum}
\usepackage{arydshln}
\usepackage{multicol}
\usepackage{titletoc}
\usepackage{mdframed}
\usepackage{tabularray}
\UseTblrLibrary{booktabs}
\usepackage[table,dvipsnames]{xcolor}
\usepackage{makecell}

\newcolumntype{L}[1]{>{\raggedright\let\newline\\\arraybackslash\hspace{0pt}}m{#1}}
\newcolumntype{R}[1]{>{\raggedleft\let\newline\\\arraybackslash\hspace{0pt}}m{#1}}

\newcommand{\ignore}[1]{}

\makeatletter
\DeclareRobustCommand\onedot{\futurelet\@let@token\@onedot}
\def\@onedot{\ifx\@let@token.\else.\null\fi\xspace}

\makeatother

\definecolor{MyBlue}{rgb}{0.46, 0.50, 0.61}
\definecolor{MyDarkBlue}{rgb}{0,0.08,0.8}
\definecolor{MyDarkGreen}{RGB}{45,155,45}
\definecolor{MyDarkRed}{rgb}{0.8,0.02,0.02}
\definecolor{MyOrange}{rgb}{1.0, 0.4, 0.2}
\definecolor{MyPurple}{RGB}{111,0,255}
\definecolor{MyRed}{rgb}{0.8,0.0,0.0}
\definecolor{MyGold}{rgb}{0.75,0.6,0.12}
\definecolor{MyDarkgray}{rgb}{0.66, 0.66, 0.66}
\definecolor{MyBrown}{rgb}{0.65, 0.16, 0.16}
\definecolor{MyMutedRose}{rgb}{0.58, 0.29, 0.35}
\definecolor{JiayuanColor}{rgb}{0.60,0.43,0.48}
\definecolor{erranColor}{rgb}{24, 40, 113}

\definecolor{citecolor}{HTML}{696FAD}

\newif\ifpropositionfirstitem
\propositionfirstitemtrue

\definecolor{bggray}{HTML}{F5F5F5}
\definecolor{pvdblue}{HTML}{DAE8FC}
\definecolor{RoseQuartzBg}{HTML}{F7CAC9}
\definecolor{RoseQuartz}{HTML}{F5A798}
\definecolor{Serenity}{HTML}{92A8D1}
\definecolor{OrangeRed}{rgb}{1.0, 0.27, 0.0}
\definecolor{RoyalBlue}{cmyk}{1, 0.50, 0, 0}
\definecolor{Turquoise}{HTML}{0F4C81}
\definecolor{mint}{rgb}{0.24, 0.71, 0.54}
\definecolor{green}{rgb}{0.0, 0.120, 0.0}

\newdimen\abovecrulesep
\newdimen\belowcrulesep
\makeatletter
\patchcmd{\@@@cmidrule}{\aboverulesep}{\abovecrulesep}{}{}
\patchcmd{\@xcmidrule}{\belowrulesep}{\belowcrulesep}{}{}
\makeatother

\definecolor{mybluetitle}{HTML}{4B527E} 

\definecolor{codegreen}{HTML}{478058}
\definecolor{codegray}{rgb}{0.5,0.5,0.5}
\definecolor{codepurple}{HTML}{4F5E80} 
\definecolor{backcolour}{rgb}{0.95,0.95,0.95}
\lstdefinestyle{mystyle}{
    backgroundcolor=\color{backcolour},
    commentstyle=\color{codegreen},
    keywordstyle=\color{magenta},
    numberstyle=\tiny\color{codegray},
    stringstyle=\color{codepurple},
    basicstyle=\ttfamily\scriptsize,
    breakatwhitespace=false,
    breaklines=true,
    captionpos=b,
    keepspaces=true,
    frame=none,
    numbersep=5pt,
    showspaces=false,
    showstringspaces=false,
    showtabs=false,
    tabsize=2
}

\newtcolorbox{promptbox}[2][]{
    enhanced, 
    breakable,
    center title,
    left*=0pt, right*=0pt,
    boxsep=2pt, left=5pt, right=5pt,
    skin first=enhanced,
    skin middle=enhanced,
    skin last=enhanced,
    colback  = backcolour,
    fonttitle=\bfseries\rmfamily,
    fontupper=\scriptsize,
    title={\footnotesize\strut{#2}},
    #1
    }

\newtcolorbox{onebox}[2][]{
    enhanced, 
    center title,
    left*=0pt, right*=0pt,
    boxsep=2pt, left=5pt, right=5pt,
    skin first=enhanced,
    skin middle=enhanced,
    skin last=enhanced,
    colframe = mybluetitle!90,
  colback  = mybluetitle!10,
    fonttitle=\bfseries\rmfamily\fontfamily{phv}\selectfont,
    title={\strut{#2}  \refstepcounter{subsubsection} \addcontentsline{toc}{subsubsection}{\string\numberline{\thesubsubsection}#2}
    },
    #1
    }

\definecolor{violet}{RGB}{111,45,168}
\annotator{mandy}{brown}
\annotator{haixin}{olive}
\annotator{wei}{blue}
\annotator{YS}{red}

\definecolor{MyGreen}{RGB}{0,128,0}
\definecolor{darkgrey}{rgb}{0.25, 0.25, 0.25}

\newcommand{\pos}[1]{\textcolor{ForestGreen}{\raisebox{-0.25ex}{\scalebox{0.75}{$_{\,+#1}$}}}}
\newcommand{\post}[1]{\textcolor{ForestGreen}{\raisebox{-0.25ex}{\scalebox{0.75}{$_{\,-#1}$}}}}
\newcommand{\posBolden}[1]{\textcolor{ForestGreen}{\raisebox{-0.25ex}{\scalebox{0.75}{$\mathbf{_{\,+#1}}$}}}}
\newcommand{\postBolden}[1]{\textcolor{ForestGreen}{\raisebox{-0.25ex}{\scalebox{0.75}{$\mathbf{_{\,-#1}}$}}}}
\newcommand{\negat}[1]{\textcolor{red!70!black}{\raisebox{-0.25ex}{\scalebox{0.75}{$_{\,+#1}$}}}}
\newcommand{\nega}[1]{\textcolor{red!70!black}{\raisebox{-0.25ex}{\scalebox{0.75}{$_{\,-#1}$}}}}

\title{HarnessBridge: Learnable Bidirectional Controller for LLM Agent Harness}

\scaiauthor{1,lead}{Xiaoxuan Wang}
\scaiauthor{1,lead}{Haixin Wang}
\scaiauthor{1}{Alexander Taylor}
\scaiauthor{1}{Jason Cong}
\scaiauthor{1}{Yizhou Sun}
\scaiauthor{1}{Wei Wang}

\affiliation{1}{University of California, Los Angeles}
\contribnote{lead}{Equal Contribution.}

\metadata[GitHub]{\url{https://github.com/mandyyyyii/HarnessBridge}}
\metadata[HuggingFace]{\url{https://huggingface.co/HarnessBridge}}
\correspondence{Haixin Wang}{whx@cs.ucla.edu}

\begin{abstract}
Large language models are increasingly deployed as agents for long-horizon tasks, yet their performance is shaped not only by model capability and environment design, but also by the harness that mediates agent--environment interaction. Existing harnesses are largely manually engineered, making them difficult to scale as trajectories grow longer and interactions become more complex. 
In this work, we ask whether harness can be generated by a learnable plug-in module that can be trained in an end-to-end fashion. 
We introduce \textbf{HarnessBridge}, a lightweight learnable harness controller that parameterizes the agent--environment interface as a bidirectional projection. 
HarnessBridge learns two bidirectional projections: observation projection, which distills raw trajectories into compact, decision-relevant states, and action projection, which converts proposed actions into executable transitions or trajectory-grounded rejections. 
We train HarnessBridge on a harness supervision dataset via unified instruction tuning. On Terminal-Bench~2.0 and SWE-bench Verified, HarnessBridge matches or surpasses strong specialized harnesses while substantially reducing token usage and trajectory length, and generalizes from smaller generators to larger commercial models. 
\end{abstract}

\begin{document}

\maketitle

\begin{figure}[!htbp]
    \centering
    \vskip-0.7em
    \includegraphics[width=.99\linewidth]{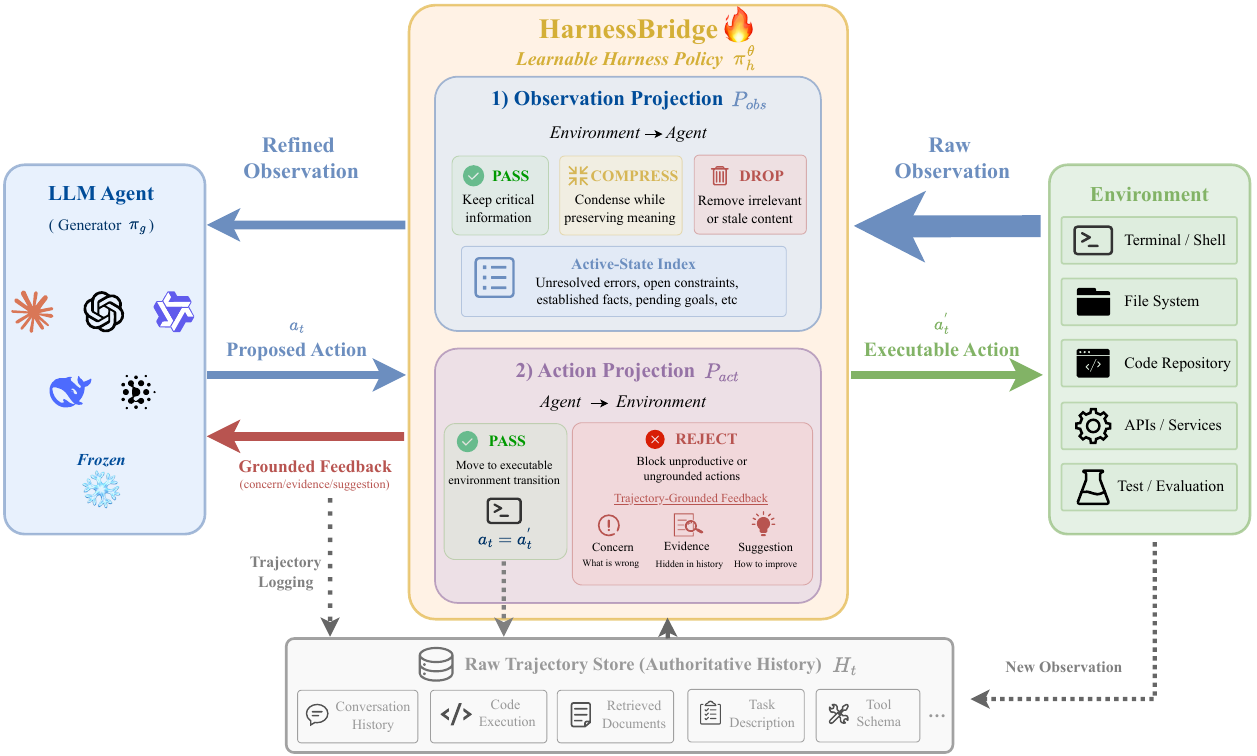}
    \caption{Overview of \textbf{HarnessBridge}, a learnable harness policy that parameterizes the agent-environment interface as a bidirectional projection: \emph{observation projection} compresses raw trajectories into a generator-visible state, while \emph{action projection} maps proposed actions to executable transitions or grounded rejections.}
    \label{fig:teaser}
    \vskip-7em
\end{figure}

\section{Introduction}

Large language models (LLMs) are increasingly deployed as agents for complex, long-horizon tasks, such as autonomous software engineering~\citep{yang2024swe,yang2026swe,merrill2026terminal,fu2026davincienvopensweenvironment}, web navigation~\citep{ning2025survey,wang2026t,wang2026arlarena}, search~\citep{wang2023dior,wang2023toward,huang2025deep}, and multi-modal tasks~\citep{wang2026vagen,chang2025survey}.
Prior progress has largely focused on two axes: stronger agents as generators and richer environments. Yet as horizons grow, a third axis becomes increasingly important: the interface that determines what information reaches the generator and what actions are committed back to the environment.

Nowadays, this interaction in the agent system is implemented by a \emph{harness}~\citep{rajasekaran2026harnessdesign,pan2026natural}: the scaffold that formats observations, manages context, invokes tools, parses outputs, validates actions, and handles environmental feedback. 
Harness engineering has become indispensable to agent performance, with many gains arising from better context construction, retry logic, summarization, and action validation. 
However, despite its importance, the harness is still usually treated as manually engineered infrastructure rather than as an optimizable policy.

Complex, manually engineered harnesses could become a central limitation in long-horizon interaction. In the \underline{environment-to-agent} direction, trajectories accumulate redundant context~\citep{anthropic2025contextengineering}, stale errors, superseded hypotheses, and low-value details, increasing token cost while obscuring decision-critical state. In the \underline{agent-to-environment} direction, model outputs may repeat ineffective actions, pursue invalidated hypotheses, enter empty loops, or issue malformed commands, consuming scarce environment steps without advancing the task~\citep{aghzal2026llm}. 

Recent work on automated or meta-level harness construction has begun to recognize harnesses as optimization targets, for example by using coding agents to improve task-specific scaffolds or search over prompt and protocol variants~\citep{lee2026meta}. These efforts show that harnesses are not fixed infrastructure, but objects that can be searched, revised, and improved. However, they typically optimize the external scaffold around an agent, rather than learning the runtime interaction policy that determines how information and actions flow between the generator and the environment. This points to a more fundamental question:

\begin{tcolorbox}[notitle, rounded corners, colframe=darkgrey, colback=white, boxrule=1.5pt, boxsep=0pt, left=0.15cm, right=0.17cm, enhanced, shadow={2.5pt}{-2.5pt}{1.5pt}{opacity=5},toprule=2pt, before skip=0.65em, after skip=0.75em]
\emph{
  {
    \centering 
  {
    \fontsize{8.5pt}{13.2pt}\selectfont 
    Can harness be formulated as an end-to-end learnable generation problem?
  }
  \\
  }
  }
\end{tcolorbox}

This reduces to optimizing the agent--environment interface: what is exposed to the agent, and what is committed to the environment.
For long-horizon behavior to remain efficient, grounded, and recoverable, this interface must preserve task-relevant state, suppress stale or redundant context, reject unproductive actions, and avoid introducing hallucinated compressed information not supported by the trajectory.

Thus, we propose \textbf{HarnessBridge}, a learnable harness policy for long-horizon LLM agents. HarnessBridge parameterizes the agent--environment interface as a bidirectional projection policy. In the environment-to-agent direction, \emph{observation projection} maps the raw interaction history into a generator-visible state, preserving decision-critical information while compressing or suppressing stale, redundant, or superseded content. In the agent-to-environment direction, \emph{action projection} maps a proposed generator action to either an executable environment transition or a trajectory-grounded rejection with feedback, preventing low-value or ungrounded actions. 

To train HarnessBridge, we construct a harness supervision dataset covering both directions of the interface. We unify these behaviors as instruction-following tasks and train a lightweight LLM via unified instruction tuning. Empirically, we evaluate HarnessBridge on long-horizon benchmarks, including Terminal-Bench~2.0 and SWE-bench Verified across a series of open-sourced and commercial models. 
HarnessBridge matches or surpasses strong specialized harnesses while substantially reducing token usage and trajectory length.
Moreover, models trained with smaller generators generalize effectively to larger commercial models. Our contributions are as fllows:

\begin{itemize}[leftmargin=*,itemsep=2pt]
    \item We introduce end-to-end harness generation for agent systems, replacing manually engineered interaction logic with a learnable harness policy.

    \item We are the first to introduce unified instruction tuning for learning bidirectional mappings between agents and environments.
    
    \item We present \textbf{HarnessBridge}, demonstrating improved efficiency and competitive or better task performance with good generalization.
\end{itemize}
\section{Related Work}

\subsection{Long-Horizon Tool-Using LLM Agents.}

LLM agents have been increasingly studied as systems that use tools to interact with external environments~\citep{wei2026agentic}. Early and representative work shows that LLMs can interleave reasoning and acting~\citep{yao2022react,AgentGen,R1-Searcher} , call external APIs~\citep{yao2024tau,barres2025tau}, and acquire reusable skills through interaction~\citep{li2026skillsbench}, enabling agents to move beyond single-turn generation toward sequential decision making~\citep{yao2023react, schick2023toolformer, wang2023voyager}. This paradigm has since been instantiated in a range of long-horizon domains, including web navigation, software engineering, terminal operation, and multi-step information seeking, with benchmarks such as WebArena, SWE-bench, and Terminal-Bench exposing realistic environments in which agents must gather information, execute actions, observe feedback, and revise their plans over many turns \citep{zhou2023tianyue, jimenez2024swebench, merrill2026terminal}. 

Despite this progress, making tool-using agents stable and scalable over long runs remains an open problem~\citep{liu2026klong}. Long-horizon interaction introduces accumulated observations, stale or redundant context, error propagation, repeated tool calls, and increasing execution cost.

\subsection{Harness Engineering for Agents.}

Existing harnesses rely on manually designed heuristics, such as trajectory summarization, retrieval-based memory, context compaction~\citep{han2025concept}, retry rules, and tool-call validation. These strategies are increasingly important for long-running agents, since merely extending the context window does not prevent trajectories from accumulating stale, redundant, or low-signal information that can degrade decision quality and distract the agent~\citep{anthropic2025harness}. However, existing methods typically implement harness behavior as static rules or separately engineered modules: they may compress history, retrieve relevant information, or validate tool calls, but they do not learn the runtime interaction policy that decides what information and actions should pass through the agent–environment interface at each step.

Recently, auto-harness begins to treat harnesses themselves as optimization targets, exploring automatic workflow optimization, prompt selection, scaffold search, or code-level harness improvement. For example, Meta-Harness~\citep{lee2026meta} represents an important step beyond manual harness engineering. Nevertheless, they typically optimize the external scaffold around an agent, rather than learning a runtime policy that continuously mediates the bidirectional flow of observations and actions during execution. In contrast, we focus on the harness-mediated interface itself.

\section{Method}
\label{sec:method}
\begin{figure*}[t]
    \centering
    \includegraphics[width=\linewidth]{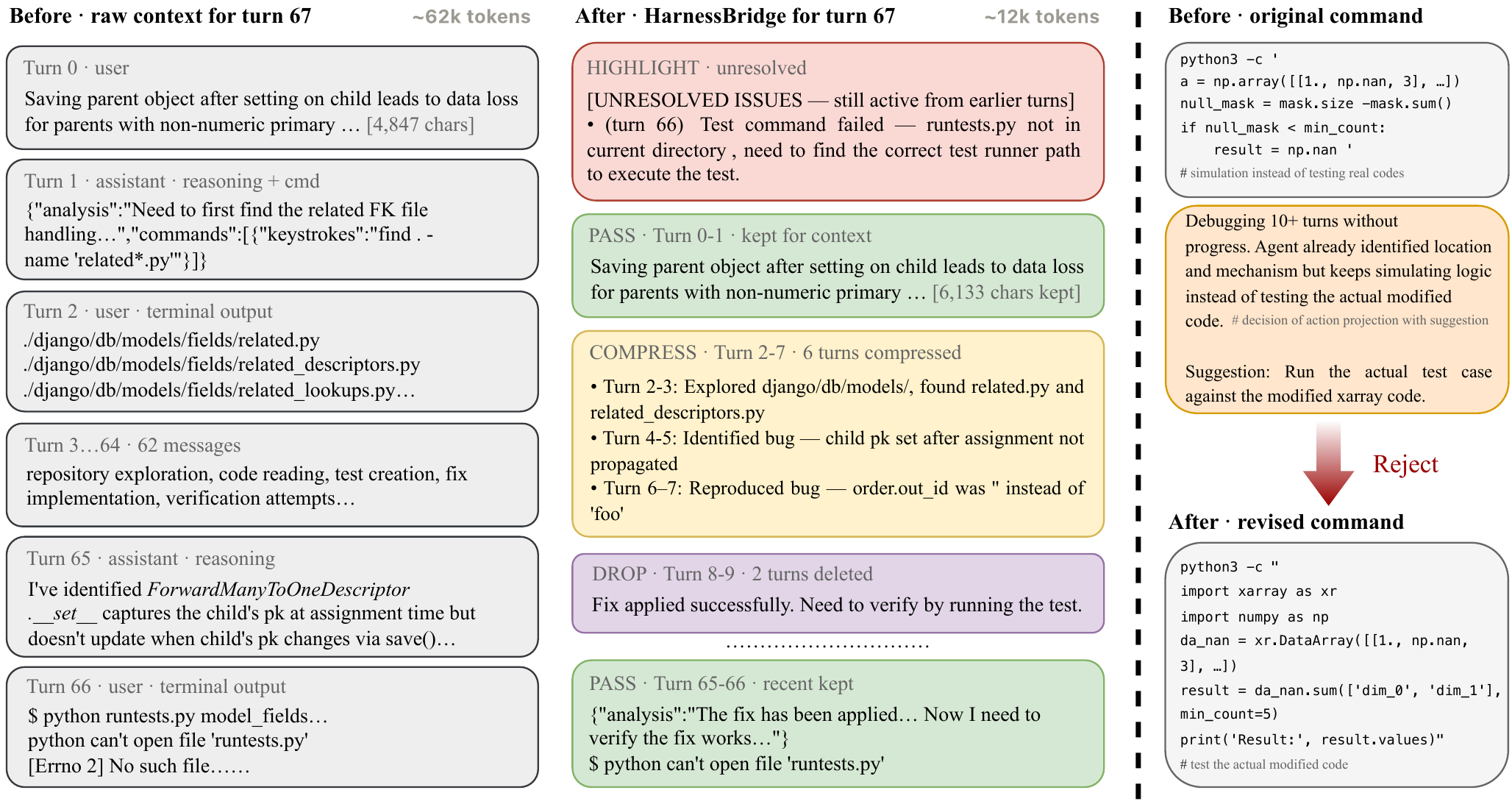}
    \caption{Examples of HarnessBridge on \texttt{django-13964\_\_dPYRYzC} (left) and \texttt{xarray-4356\_\_fmRfApG} (right) from SWE-Bench Verified, illustrating observation projection and action projection, respectively.}
    \label{fig:example}
    \vspace{-0.3cm}
\end{figure*}

We consider a tool-using LLM agent interacting with an environment
$\mathcal{E}$ through a tool interface $\mathcal{T}$. At turn $t$, the
interaction history is denoted by $H_t$, which contains previous model
outputs, tool calls, environment observations, retrieved evidence, code
snippets, execution results, and scaffold-provided reminders. 
Given a system prompt $s$ and task instruction $q$, a conventional harness
serializes this history into the generator input and samples an action
from a generator policy:
\begin{equation}
    a_t \sim \pi_g(\cdot \mid s,q,H_t)
\end{equation}
The action is then executed in the environment, producing an observation
$o_t$, and the history is updated as
$H_{t+1} = H_t \cup \{a_t,o_t\}$.

This conventional formulation treats the harness as a fixed interface
between the generator and the environment. Yet in long-horizon tool-use,
this interface plays a substantive role in shaping agent behavior. The
generator conditions on a harness-constructed representation of the
trajectory rather than the raw environment state, while the environment is
affected only through actions dispatched by the harness. Consequently,
the harness implicitly defines both the generator's effective observation and the action exposed to the environment.

\subsection{HarnessBridge}
\label{sec:formulation}


We formulate harness engineering as an end-to-end learnable interaction problem, where a harness learns to mediate what the agent observes and what the environment executes. Specifically, HarnessBridge parameterizes the harness as a learnable bidirectional interaction policy $\pi_h$:
\begin{equation}
    \pi_h : (s,q,H_t,a_t) \mapsto (\widetilde{H}_t,a_t')
\end{equation}
where $\widetilde{H}_t$ is the state exposed to the generator and $a_t'$
is the action exposed to the environment. The generator policy $\pi_g$
remains fixed; optimization is performed only over the harness policy
$\pi_h$. 

We instantiate $\pi_h$ as two directional projections over the
agent--environment interface. In the environment-to-agent direction,
the observation projection maps the raw history to a generator-visible
state:
\begin{equation}
     \widetilde{H}_t = P_{\mathrm{obs}}(s,q,H_t)
\end{equation}
The generator then samples an action
$a_t \sim \pi_g(\cdot \mid s,q,\widetilde{H}_t)$. In the
agent-to-environment direction, the action projection maps the
proposed action to the environment-facing action:
\begin{equation}
    a_t' = P_{\mathrm{act}}(s,q,H_t,a_t)
\end{equation}
Only $a_t'$ is exposed to the environment, inducing the transition
$o_t=\mathcal{E}(a_t')$ and the updated history
$H_{t+1}=H_t\cup\{a_t',o_t\}$.
Together, $P_{\mathrm{obs}}$ and $P_{\mathrm{act}}$ define a learned
interface between the generator and the environment, controlling both the
state exposed to the generator and the action exposed to the environment.

\subsubsection{Observation Projection}
\label{sec:observation_projection}

Observation projection specifies how the raw interaction history is exposed
to the generator. Let the history at turn $t$ be represented as a sequence
of interaction units $H_t=(h_1,\ldots,h_t)$, where each $h_i$ may denote a
full interaction turn or a finer-grained component, such as an action,
observation, tool feedback, intermediate artifact, retrieved evidence, or
other state produced during the interaction. Observation projection learns
how these units should be represented in the generator-visible state. 

We expand the observation projection as
\begin{equation}
    \widetilde{H}_t
    =
    P_{\mathrm{obs}}(s,q,H_t)
    =
    (U_t,\widetilde{h}_1,\ldots,\widetilde{h}_t)
\end{equation}
where $U_t$ is an active-state index placed before the projected history,
and $\widetilde{h}_i$ is the projected representation of history unit
$h_i$. Both $U_t$ and $\{\widetilde{h}_i\}_{i=1}^t$ are induced by the
learned observation policy. Specifically, the policy predicts:
\begin{equation}
\begin{aligned}
    U_t &= P_{\mathrm{obs}}^{\mathrm{state}}(s,q,H_t), \\
    z_i &= P_{\mathrm{obs}}^{\mathrm{exp}}(s,q,H_t,h_i),
\end{aligned}
\end{equation}
where $z_i \in \{\textsc{Pass}, \textsc{Compress}, \textsc{Drop}\}$. The exposure decision $z_i$ determines the projected form of each history
unit:
\begin{equation}
    \widetilde{h}_i =
    \begin{cases}
        h_i, & z_i=\textsc{PASS},\\
        \mathrm{Compress}(h_i), & z_i=\textsc{Compress},\\
        \varnothing, & z_i=\textsc{Drop}.
    \end{cases}
\end{equation}

The active-state index $U_t$ is a learned view extracted from the raw
history. It records information that should remain immediately visible to
the generator, including unresolved errors, open constraints, established
facts, pending goals, and remaining decision variables. By placing $U_t$
before the projected chronological history, HarnessBridge makes the current
interaction state explicit without requiring the generator to reconstruct
it from a long trajectory.

Observation projection acts as a learned exposure function over the
interaction history. It preserves decision-critical units, compresses
relevant but verbose units, and suppresses units whose information is
irrelevant, redundant, or superseded. 
\subsubsection{Action Projection}
\label{sec:action_projection}

Action projection specifies how generator proposals are exposed to the
environment. In long-horizon interaction, syntactically valid actions may
still be redundant, weakly grounded, inconsistent with accumulated
evidence, or unlikely to advance the task, wasting interaction
steps and time. Action projection therefore learns an environment-facing
map from proposed actions to executable or rejected transitions,
conditioned on the current task state and interaction history.

Given the raw interaction history $H_t$ and a proposed action $a_t$,
$P_{\mathrm{act}}$ predicts
\begin{equation}
    P_{\mathrm{act}}(s,q,H_t,a_t) = (d_t,\rho_t)
\end{equation}
where
\[
    d_t \in \{\textsc{Pass},\textsc{Reject}\}.
\]
When $d_t=\textsc{Pass}$, the projected action is the original proposal,
$a_t'=a_t$, and the environment transition proceeds. When
$d_t=\textsc{Reject}$, no environment step is taken; the projected action
is null, $a_t'=\varnothing$, and the feedback $\rho_t$ is returned to the
generator as part of the subsequent projected state.

The feedback $\rho_t$ is required to be grounded in the current
interaction history. We represent it as
\[
    \rho_t =
    (\textit{concern},\textit{evidence},\textit{suggestion}),
\]
where the \textit{concern} states why the proposed action is unlikely to
be a productive transition, the \textit{evidence} identifies the specific
trajectory information supporting this assessment, and the
\textit{suggestion} provides an actionable direction for revision. If
$P_{\mathrm{act}}$ cannot provide trajectory-grounded evidence, it
defaults to \textsc{Pass}.

Thus, action projection serves as an environment-facing interface policy:
it determines which proposed transitions should affect the environment
under the current task state and interaction budget. By requiring
trajectory-grounded evidence for rejection, $P_{\mathrm{act}}$ reduces
unproductive environment steps while preserving informative exploration.

\subsection{Unified Instruction Fine-tuning}
\label{sec:training}

\noindent\textbf{Training Formulation.} We formulate learning the bidirectional interface as a unified conditional generation problem. Rather than training separate modules for observation and action projection, we parameterize both $P_{\mathrm{obs}}$ and $P_{\mathrm{act}}$ with a shared policy $P_{\theta}$. Given an instruction specifying the projection objective, together with the task specification and current trajectory, $P_{\theta}$ is trained to generate the corresponding trajectory-grounded interface transformation. Under this formulation, observation projection and action projection differ only in their instruction and target format: the former produces a generator-visible state, while the latter produces an environment-facing pass/reject decision with grounded feedback when needed.

\noindent\textbf{Data Curation.} To construct supervised data, we instantiate observation and action projections with prompted instruction-tuned models under multiple intervention regimes, producing traces that include raw trajectories, projected states, proposed actions, projection decisions, feedback, and environment outcomes. We then filter these traces for high-quality supervision by retaining only successful trajectories and using an LLM judge to assess projection quality. Observation-projection examples are selected for schema consistency, faithful compression, and preservation of decision-critical evidence, while action-projection examples are retained when pass/reject decisions and rejection feedback are trajectory-grounded and actionable. More details are shown in Appendix \ref{sec:appen_curation}.

\noindent\textbf{Raw Trajectory Preservation.} HarnessBridge does not destructively overwrite the interaction history. 
The raw trajectory $H_t$ is always retained as the authoritative record, while $P_{\theta}$ only decides what projected view $\widetilde{H}_t$ should be exposed to the generator at each turn. Thus, compression is triggered selectively rather than applied as an irreversible update after every interaction step. 
This design mitigates two common risks of trajectory compression: hallucinated summaries and the accidental removal of details needed for later reasoning. 
When compression is used, we require the projected state to be provenance-aware: compressed statements, unresolved errors, constraints, and task-relevant facts must be grounded in specific spans of the original trajectory. 

\section{Experiments}
\label{sec:exp}

\begin{table*}[t]
\vspace{0.8cm}
\centering
\small
\setlength{\tabcolsep}{5pt}
\renewcommand{\arraystretch}{1.15}
\begin{tabular}{@{}l cc cc cc cc@{}}
\toprule
\multirow{3}{*}{\textbf{Harness}} & \multicolumn{4}{c}{\textbf{Qwen3.5-35B-A3B}} & \multicolumn{4}{c}{\textbf{GLM-4.7-Flash}} \\
\cmidrule(lr){2-5} \cmidrule(lr){6-9}
& \multicolumn{2}{c}{Terminal-Bench 2.0} & \multicolumn{2}{c}{SWE-Bench Verified} & \multicolumn{2}{c}{Terminal-Bench 2.0} & \multicolumn{2}{c}{SWE-Bench Verified} \\
\cmidrule(lr){2-3} \cmidrule(lr){4-5} \cmidrule(lr){6-7} \cmidrule(lr){8-9}
& SR $\uparrow$ & Token $\downarrow$ & SR $\uparrow$  & Token $\downarrow$& SR $\uparrow$  & Token $\downarrow$& SR $\uparrow$  & Token $\downarrow$\\
\midrule
\multicolumn{9}{c}{\textit{Manual Harness}} \\
\midrule
Terminus~2
& 30.3 & 2.31
& \textbf{61.6} & 1.47
& 19.1 & 1.87
& 45.2 & 1.51 \\

Terminus-KIRA
& 27.0\nega{10.9\%} & 9.59\negat{315.2\%}
& 46.0\nega{25.3\%} & 9.77\negat{564.6\%}
& 6.7\nega{64.9\%} & 4.90\negat{162.0\%}
& 37.8\nega{16.4\%} & 6.78\negat{349.0\%} \\

mini-SWE-agent
& 29.2\nega{3.6\%} & 6.32\negat{173.6\%}
& 59.8\nega{2.9\%} & 5.92\negat{302.7\%}
& 11.2\nega{41.4\%} & 1.60\post{14.4\%}
& 45.4\nega{0.4\%} & 3.91\negat{158.9\%} \\

OpenHands
& 27.0\nega{10.9\%} & 2.61\negat{13.0\%}
& 52.6\nega{14.6\%} & 1.96\negat{33.3\%}
& 13.5\nega{29.3\%} & 1.28\post{31.6\%}
& 42.0\nega{7.1\%} & 3.69\negat{144.4\%} \\

Qwen-Coder
& 24.7\nega{18.5\%} & 4.19\negat{81.4\%}
& 58.8\nega{4.5\%} & 3.86\negat{162.6\%}
& 10.1\nega{47.1\%} & 1.41\post{24.6\%}
& 42.0\nega{7.1\%} & 4.34\negat{187.4\%} \\
\midrule
\multicolumn{9}{c}{\textit{Auto-Harness}} \\
\midrule
Meta-Harness
& 31.5\pos{4.0\%} & 2.20\post{4.8\%}
& 59.2\nega{3.9\%} & 1.92\negat{30.6\%}
& 14.6\nega{23.6\%} & 1.33\post{28.9\%}
& \textbf{49.6\posBolden{9.7\%}} & 2.43\negat{60.9\%} \\

\textsc{HarnessBridge}
& \textbf{33.7\posBolden{11.2\%}} & \textbf{1.23\postBolden{46.8\%}}
& 60.2\nega{2.3\%} & \textbf{1.13\postBolden{23.1\%}}
& \textbf{20.2\posBolden{5.8\%}} & \textbf{0.42\postBolden{77.5\%}}
& 46.0\pos{1.8\%} & \textbf{1.48\postBolden{2.0\%}} \\
\bottomrule
\end{tabular}
\caption{Success rate (SR) and average token usage in millions across harnesses on Terminal-Bench~2.0 and SWE-bench Verified. Subscripts denote relative changes against Terminus~2. For SR, \textcolor{ForestGreen}{green} indicates improvement and \textcolor{red!70!black}{red} indicates degradation. For token usage, \textcolor{ForestGreen}{green} indicates reduction and \textcolor{red!70!black}{red} indicates increase. Best results are highlighted in bold.}
\label{tab:main}
\vspace{-0.3cm}
\end{table*}

Our experiments are designed to answer the following three questions. 
\textbf{Q1:} Can HarnessBridge effectively reduce the token consumption of agents while maintaining strong task performance? 
\textbf{Q2:} Can HarnessBridge, instruction-tuned on Qwen3.5, generalize to larger commercial models and different environments? 
\textbf{Q3:} Do the bidirectional mappings in HarnessBridge contribute effectively to agent--environment interaction?

\subsection{Experiment Setup}
\label{sec:exp-setup}
\paragraph{Benchmarks and Baselines.}
We focus on coding as a representative class of tool-intensive, long-horizon agent tasks, evaluating on Terminal-Bench~2.0~\citep{merrill2026terminal} and SWE-bench Verified~\citep{jimenez2024swebench}. We report the success rate and average input-token consumption for the generator.
The baseline comparison includes representative scaffolds from several categories:
\textit{Terminal-Bench scaffolds}: Terminus~2~\citep{merrill2026terminal} (the official TB-2.0 reference scaffold) and Terminus-KIRA~\citep{terminuskira2026};
\textit{SWE-Bench scaffolds}: mini-swe-agent~\citep{yang2024swe};
\textit{production harnesses}: OpenHands~\citep{wang2024openhands} and Qwen-Coder;
and \textit{Automatic scaffold optimization}: Meta-Harness~\citep{lee2026meta}, the closest prior work that explicitly searches over scaffold behavior. 

\paragraph{Model Backbones.}
Experiments use seven frozen generator models with different parameter sizes: Qwen3.5-35B-A3B~\citep{qwen3.5}, GLM4.7-Flash~\citep{glm4_7}, DeepSeek-V4-Flash, DeepSeek-V4-Pro~\citep{deepseek_v4}, GPT-5.4-nano, GPT-5.4~\citep{openai_gpt5_4}, and Claude-Opus-4.7. 
Qwen3.5-35B-A3B and GLM4.7-Flash are open-sourced and served with SGLang~\citep{zheng2024sglang} at temperature $0.6$, while the remaining models are accessed via commercial APIs.
HarnessBridge is initialized from the lightweight Qwen3.5-0.8B~\citep{qwen3.5} and instruction-tuned to support harness-control decisions. Additional details are provided in Appendix~\ref{sec:appen_curation}.

\subsection{Main Results}
\label{sec:exp-main}

The experimental results provide a clear affirmative answer to \textbf{Q1}. Table~\ref{tab:main} reports success rate (SR) and average input-token consumption across harnesses, generators, and benchmarks. On SWE-bench Verified with Qwen3.5-35B-A3B, \textsc{HarnessBridge} remains competitive while using the lowest token budget among reported harnesses. Overall, \textsc{HarnessBridge} jointly improves success rate and reduces token consumption compared to other harness design.

It is worth noting that HarnessBridge is instruction-tuned only on trajectories from SWE-bench. Therefore, Terminal-Bench~2.0 can be viewed as an out-of-domain environment. Nevertheless, HarnessBridge still achieves strong performance, suggesting that the learned harness policy generalizes beyond the training environment. On Terminal-Bench~2.0, \textsc{HarnessBridge} achieves the highest success rate under both generators, reaching $33.7\%$ with Qwen3.5-35B-A3B and $20.7\%$ with GLM-4.7-Flash. Compared with Terminus~2, this corresponds to success rate gains of $+11.2$ and $+8.4$ percentage points, while reducing average token consumption by $42.4\%$ and $77.0\%$, respectively.

\subsection{Generalization on Commercial Models}
\textbf{Q2} is to answer whether a harness supervised from traces of a single generator transfers to generators with different inference-time behaviors. 
To evaluate this, we test \textsc{HarnessBridge} on five unseen generators spanning multiple model families and capability levels (Table~\ref{tab:agent_harness_results}).
Across all settings, \textsc{HarnessBridge} preserves or improves success rate while reducing token consumption. The largest improvement is observed on GPT-5.4-Nano, where success rate increases from $15.7\%$ to $22.5\%$ and average token usage decreases from $9.77$M to $0.91$M. On GPT-5.4, \textsc{HarnessBridge} maintains the same success rate of $53.9\%$ while reducing token usage by approximately $89\%$. On DeepSeek-V4-Pro, where the baseline is already efficient, it matches the baseline success rate of $57.3\%$ with slightly lower token cost. On Claude-Opus-4.7, HarnessBridge maintains strong task performance while further improving efficiency. It increases success rate from $64.0\%$ to $65.2\%$ and reduces token usage by $26.9\%$.

These results suggest that \textsc{HarnessBridge} captures interaction-level control patterns that transfer beyond the generator used for supervision. The gains are larger when the baseline incurs high interaction cost, and smaller when the baseline is already efficient.
\begin{table*}[t]
\centering
\small
\setlength{\tabcolsep}{1.7pt}
\renewcommand{\arraystretch}{1.12}
\begin{tabular}{lcc cc cc cc cc}
\toprule
\multirow{2}{*}{\textbf{Harness}}
& \multicolumn{2}{c}{\textbf{GPT-5.4-Nano}}
& \multicolumn{2}{c}{\textbf{GPT-5.4}}
& \multicolumn{2}{c}{\textbf{DeepSeek-V4-Flash}}
& \multicolumn{2}{c}{\textbf{DeepSeek-V4-Pro}}
& \multicolumn{2}{c}{\textbf{Claude-Opus-4.7}} \\
\cmidrule(lr){2-3}
\cmidrule(lr){4-5}
\cmidrule(lr){6-7}
\cmidrule(lr){8-9}
\cmidrule(lr){10-11}
& SR $\uparrow$& Token $\downarrow$
& SR $\uparrow$& Token $\downarrow$
& SR $\uparrow$& Token $\downarrow$
& SR $\uparrow$& Token $\downarrow$
& SR $\uparrow$& Token $\downarrow$ \\
\midrule
Terminus~2
& 18.0 & 9.80
& 53.9 & 9.41
& 49.4 & 2.18
& 57.3 & 1.02
& 64.0 & 0.26 \\
\textsc{HarnessBridge}
& 22.5\pos{25.0\%} & 0.91\post{90.7\%}
& 53.9 & 0.99\post{89.5\%}
& 53.9\pos{9.1\%} & 1.22\post{44.0\%}
& 57.3 & 0.94\post{7.8\%}
& 65.2\pos{1.9\%} & 0.19\post{26.9\%} \\
\bottomrule
\end{tabular}
\caption{Success rate (SR) and average input token usage (in millions) on Terminal-Bench~2.0 across models.}
\label{tab:agent_harness_results}
\vspace{-0.5cm}
\end{table*}
\begin{figure}[h]
  \centering
  \includegraphics[width=0.6\linewidth]{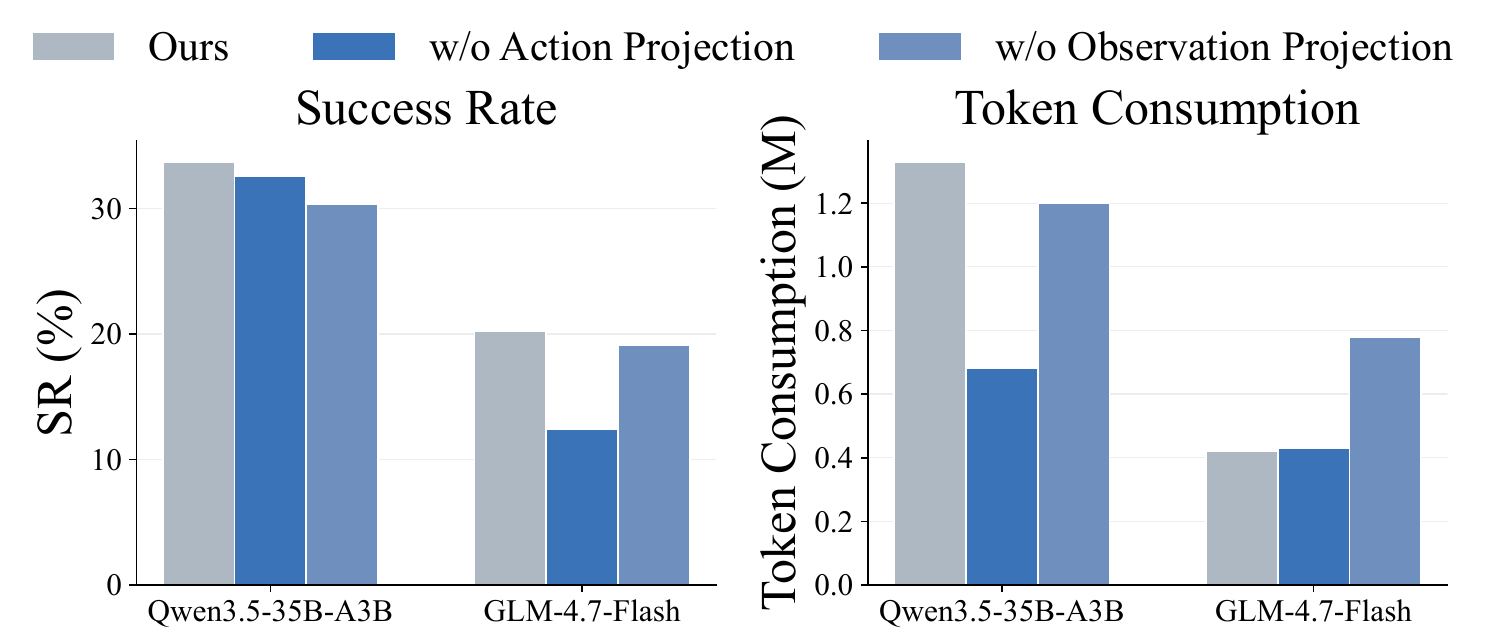}
  \caption{Ablation study of objection and action projections on Qwen3.5-35B-A3B and GLM-4.7-Flash.}
  \label{fig:ablation}
\end{figure}

\subsection{Robustness to the Supervision Source}
A natural concern is whether harness-control supervision must be sampled
from the same generator it will later control. We test this by
reconstructing the full pipeline with DeepSeek-V4 as the sampling model in
place of Qwen3.5-35B-A3B, yielding \textsc{HarnessBridge-D}. Despite the
change in data source, \textsc{HarnessBridge-D} improves task success over
the Terminus~2 baseline across heterogeneous generators while substantially
reducing token usage---including the GPT-5.4 family, which shares no lineage
with the sampling model (e.g., $18.0 \rightarrow 23.6$ success rate with a $\sim$90\%
token reduction on GPT-5.4-Nano, and $57.3 \rightarrow 59.6$ on
DeepSeek-V4-Pro). This indicates that the learned interface policy captures
general properties of effective harness control rather than artifacts of a
particular generator's trajectories. Full curation details and per-model
results appear in Appendix~\ref{sec:appen_curation_d} and~\ref{sec:appen_result_d}.

\subsection{Ablation Study}
\label{sec:exp-ablation}
We ablate the action projection and observation projection modules on Terminal-Bench~2.0 to assess their individual contributions. Figure~\ref{fig:ablation} reports success rate and token consumption across two backbone models. Removing either component reduces success rate on both backbones, indicating that both projections contribute to task performance. Action projection regularizes the model’s outputs toward executable and task-relevant operations, while observation projection organizes environment feedback into a more compact and informative form for downstream reasoning. While some ablated variants achieve lower token usage, this comes at a consistent cost in success rate, suggesting that the projections improve task completion rather than merely compressing context. Together, the results support the inclusion of both modules.
\vspace{-0.2cm}
\section{Analysis}
\vspace{-0.2cm}

\subsection{Trajectory Case Studies}
\label{sec:exp-case}
Figure~\ref{fig:example} illustrates the two projection mechanisms of \textsc{HarnessBridge} using examples from SWE-bench Verified. In the \texttt{django-13964} trajectory on the left, the agent has accumulated a long history of repository exploration, code inspection, and test attempts by turn 67. The raw context is dominated by intermediate steps that are no longer directly relevant to the next action. Observation projection reduces this context by preserving the initial task description and the most recent turns, summarizing earlier exploratory steps, and omitting redundant turns. Importantly, the failed test invocation from the most recent context is retained as an active-state item, ensuring that the current blocker remains visible after compression. This example shows that observation projection can reduce context noise while preserving decision-relevant state.

On the right, the \texttt{xarray-4356} trajectory illustrates a failure mode addressed by action projection. Although the agent has identified the relevant code path and likely bug mechanism, it repeatedly performs indirect checks rather than testing the modified codebase. \textsc{HarnessBridge} rejects the next redundant check and provides a concrete redirect to run the actual test against the patched code. The agent's next action follows this suggestion and produces a meaningful signal about the patch state. This example shows that action projection can interrupt redundant verification loops and redirect the agent toward more informative environment feedback without modifying the underlying generator.

\begin{figure}[t]
  \centering
  \includegraphics[width=0.6\linewidth]{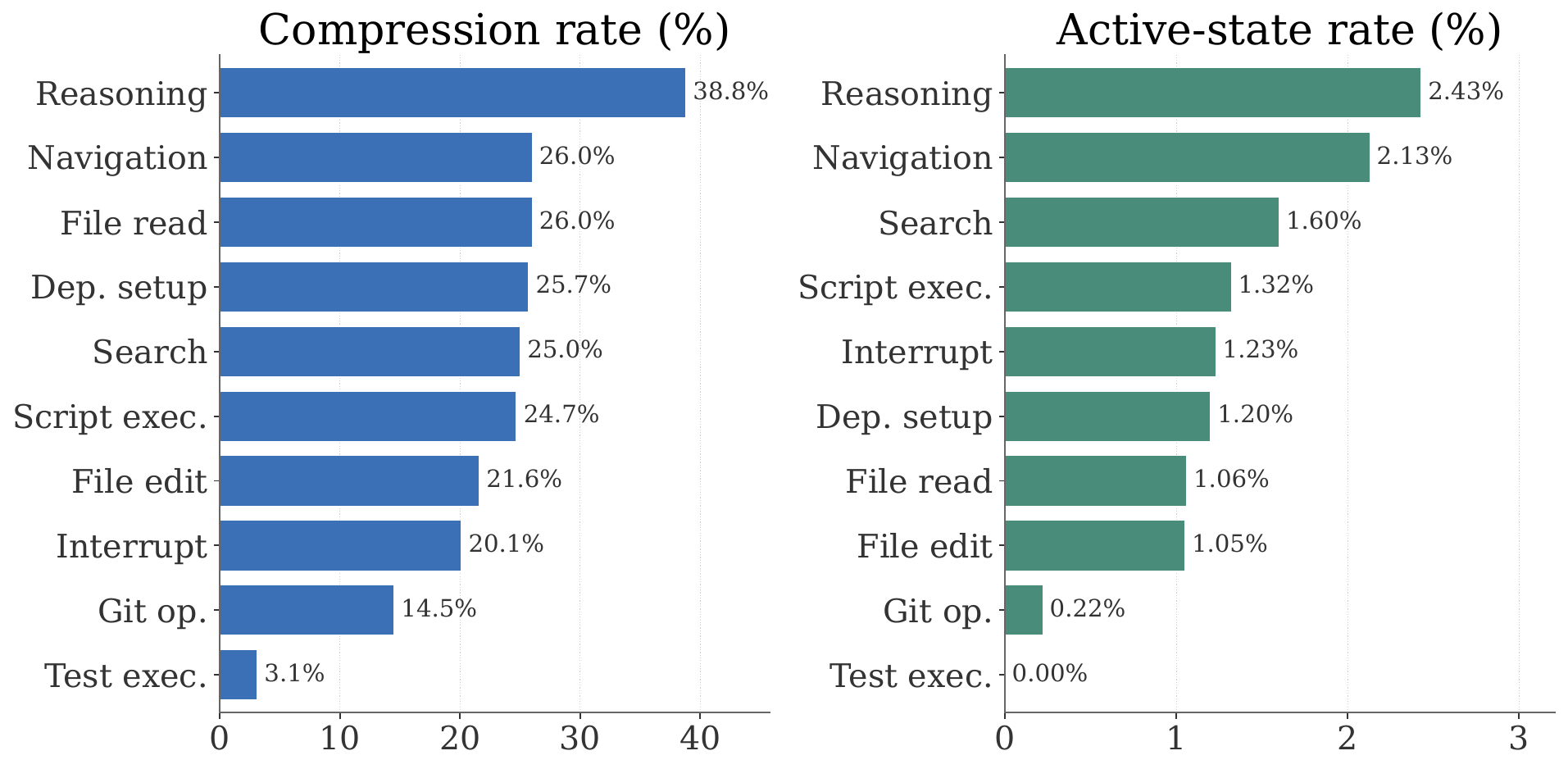}
  \caption{Observation projection behavior by action category.
    \emph{Left}: compression rate, the per-turn fraction of downstream
    invocations in which the projection compressed the turn.
    \emph{Right}: active-state rate, the per-turn fraction of
    invocations that lifted content from the turn into the persistent
    active-state block. Both metrics are means of per-turn rates over
    all turns in each category.}
  \label{fig:projection-rates}
\end{figure}
\subsection{Category-Level Analysis}
\label{sec:per-category}

To understand \emph{what} HarnessBridge has learned, we classify every
agent turn into one of ten mutually-exclusive action categories defined
by syntactic patterns over the action's command field
(e.g.\ \texttt{file\_edit}, \texttt{script\_execution},
\texttt{reasoning\_only}; refer to
Appendix~\ref{app:category-analysis}) and measure how HarnessBridge treats each category.

Figure~\ref{fig:projection-rates} reports two per-turn lifecycle
averages by category. Both metrics share the same construction: for
each turn $t$, we record the projection's decision at every
downstream invocation that re-evaluated the trajectory, compute a
single ratio per turn, and then average those ratios over all turns
in the category. The \emph{compression rate} of a turn is the
fraction of its downstream invocations in which the projection
compressed that turn; the \emph{active-state rate} of a turn is the
fraction of its downstream invocations in which the projection
promoted that turn into the persistent active-state block. Two patterns are immediately visible.
First, compression is highly selective: rates range from $3.1\%$ on \texttt{test\_execution} --- the category most likely to carry the decisive verification signal --- to $38.8\%$ on \texttt{reasoning\_only}. Categories
that are information-dense but quickly stale (reasoning, navigation,
search, file reads) are compressed at roughly $20$--$40\%$, while
content with longer downstream utility (test output, git diffs) is
largely preserved. Second, the active-state ranking tracks the
compression ranking closely (\texttt{reasoning\_only},
\texttt{navigation}, \texttt{search} top both panels): the projection
is \emph{not} simply discarding what it compresses, but distilling key
facts from those turns into a persistent slot. The two mechanisms
operate in tandem. 
\subsection{Outcome-Level Efficiency Analysis}
\begin{figure}[t]
  \centering
  \includegraphics[width=0.6\linewidth]{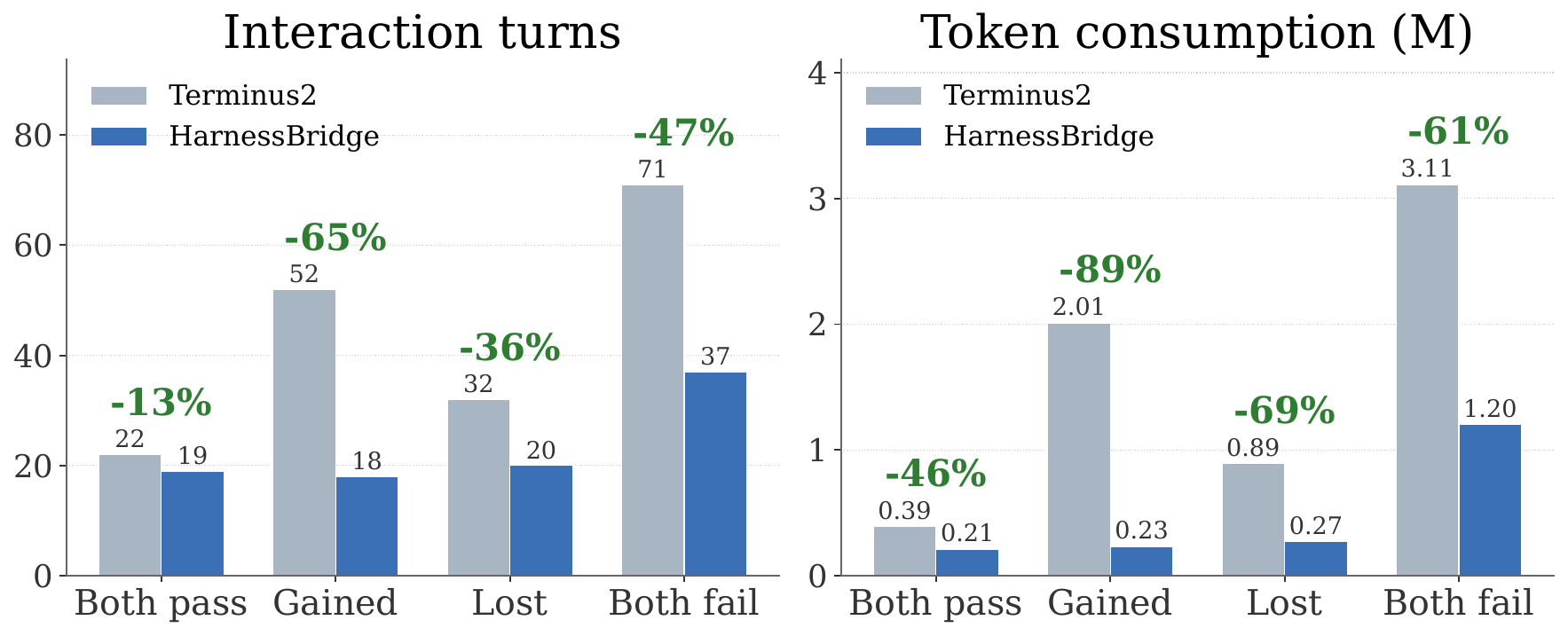}
  \caption{Outcome-level comparison of interaction turns and token consumption between Terminus2 and HarnessBridge across task success categories. }
  \label{fig:outcome}
\end{figure}
We evaluate HarnessBridge against the baseline across 178 tasks using Qwen3.5-35B-A3B and GLM4.7-Flash generators, recording pass/fail, turn count, and token consumption per run. Tasks are partitioned by joint outcome: \emph{Both passed}, \emph{Gained} (HarnessBridge only), \emph{Lost} (baseline only), and \emph{Both fail}. Figure~\ref{fig:outcome} reports per-category task counts and mean turns and tokens for each harness, with relative deltas.
\paragraph{Findings.}
HarnessBridge reduces both turns and token consumption across all outcome
categories. In each category, the reduction in tokens is larger than the
reduction in turns, suggesting that HarnessBridge not only shortens
trajectories but also produces more compact per-turn context. This
efficiency gain is observed regardless of whether the task ultimately
succeeds.

The \emph{Gained} category shows the largest difference. On tasks
where HarnessBridge succeeds and the baseline fails, HarnessBridge reaches
a passing solution in 18 turns on average, compared with 52 turns for the
baseline ($-65\%$), while using only 11\% of the baseline token budget
($-89\%$). This pattern suggests that the baseline often spends many turns
on unproductive exploration, whereas HarnessBridge can converge earlier by
maintaining a more compact and decision-relevant interaction history.
Thus, the observed efficiency improvements are associated not only with
lower cost, but also with improved task outcomes in a subset of cases.

\section{Conclusion}
HarnessBridge recasts the agent–environment interface as a learnable harness policy. Through jointly trained bidirectional projections, it compresses raw trajectories into decision-critical agent context and maps proposed actions into executable transitions or trajectory-grounded rejections. HarnessBridge achieves competitive or stronger performance with lower token cost and shorter trajectories on the benchmarks, while transferring from small to larger commercial models.

\bibliographystyle{plainnat}
\bibliography{main}

\clearpage
\appendix

\section{Preliminary Study}
\label{appendix:implementation}

\paragraph{Harness Backbone Comparison.}
\begin{table}[h]
\centering
\small
\setlength{\tabcolsep}{4pt}
\begin{tabular}{lccc}
\toprule
Model & Benchmark & SR (\%) & Tokens (M) \\
\midrule
Qwen3.5-0.8B & SWE & 58.3 & 1.43 \\
Qwen3.5-0.8B & TB  & 24.7 & 1.99 \\
Qwen3.5-35B-A3B  & SWE & 59.1 & 1.38 \\
Qwen3.5-35B-A3B  & TB  & 29.2 & 0.53 \\
\bottomrule
\end{tabular}
\caption{Evaluation results across vanilla Qwen3.5-0.8B and Qwen3.5-35B-A3B models.}
\label{tab:results}
\end{table}
We compare different backbone models for the harness component, including vanilla Qwen3.5-35B-A3B and vanilla Qwen3.5-0.8B. The results show that the fine-tuned Qwen3.5-0.8B HarnessBridge model achieves performance and efficiency comparable to the much larger Qwen3.5-35B-A3B harness, while requiring substantially lower inference cost. This suggests that harness-control behavior can be effectively distilled into a small model.

\paragraph{Effect of Reserved History Window on Observation Projection.}
\begin{table}[h]
\centering
\small
\setlength{\tabcolsep}{4pt}
\begin{tabular}{lcc}
\toprule
Reserved Turns & SR (\%) & Tokens (M) \\
\midrule
N10  & 56.9 & 0.72 \\
N20  & 56.4 & 0.65 \\
N30  & 60.4 & 0.74 \\
N50  & 61.3 & 0.96 \\
N70  & 58.8 & 1.08 \\
N100 & 58.0 & 1.40 \\
\bottomrule
\end{tabular}
\caption{Effect of the number of reserved turns on SWE-Verified.}
\label{tab:reserved-turns}
\end{table}
We study the effect of the reserved history window $N$ on SWE-bench Verified. The reserved window controls when observation projection is activated: before this window, the full recent history is preserved; beyond it, compression and dropping are enabled. The results indicate that both overly aggressive and overly conservative compression can hurt performance or efficiency. A moderate history window achieves a better trade-off, suggesting the need for dynamically learned context management rather than a fixed compression rule.

\paragraph{Effect of rejection mode on Action Projection.}
\begin{table}[h]
\centering
\small
\setlength{\tabcolsep}{4pt}
\begin{tabular}{llcc}
\toprule
Benchmark & Mode & SR (\%) & Tokens (M) \\
\midrule
SWE & Rules-only  & 60.9 & 1.74 \\
SWE & Tolerant    & 61.5 & 1.50 \\
SWE & Strict      & 56.7 & 2.08 \\
TB  & Rules-only  & 29.2 & 4.25 \\
TB  & Tolerant    & 30.3 & 2.34 \\
TB  & Strict      & 25.8 & 2.15 \\
\bottomrule
\end{tabular}
\caption{Action projection under different rejection modes: rules-only, tolerant, and strict.}
\label{tab:harness-ablation}
\end{table}
We evaluate action projection under different rejection modes. In the rule-only mode, the harness rejects only simple invalid actions, such as malformed syntax, empty commands, duplicated commands, or repeated command patterns, and returns predefined feedback asking the generator to revise its action. However, rule-based rejection alone has limited impact, since many inefficient actions are syntactically valid but semantically unproductive. In the learned rejection setting, strict rejection can over-intervene and reject otherwise useful actions, reducing task success. A more tolerant rejection mode achieves a better balance between blocking unproductive environment steps and preserving valid agent behavior. These results highlight the importance of calibrating action projection to avoid both under-intervention and excessive rejection.

\section{Action Category Analysis }
\label{app:category-analysis}

We partition agent turns into ten mutually exclusive categories
(Table~\ref{tab:category-rules}), defined by syntactic patterns over
the action's command field. We chose these categories to (a)~cover the
dominant agent behaviors observed on SWE-bench Verified and Terminal
Bench while (b)~being deterministically derivable from command syntax,
so the classifier is reproducible and does not require semantic
inspection of the observation. 

Intuitively, the ten categories cover three behavioral modes:
\emph{information gathering} (\texttt{file\_read}, \texttt{search},
\texttt{navigation}, \texttt{git\_operation}),
\emph{state-changing action} (\texttt{file\_edit},
\texttt{dependency\_setup}, \texttt{interrupt\_abort}), and
\emph{execution/verification} (\texttt{script\_execution},
\texttt{test\_execution}). \texttt{reasoning\_only} is the residual
class for turns whose action carries no parseable command,
characteristic of larger reasoning-tuned generators. Submission turns
(\texttt{COMPLETE\_TASK\_AND\_SUBMIT}, \texttt{mark\_task\_complete})
are detected by the classifier but excluded from analysis: a
submission is always the final turn of its trajectory and therefore
has no downstream projection invocations against which to measure a
lifecycle.

\begin{table*}[h]
\centering
\small
\begin{tabular}{p{0.20\linewidth} p{0.55\linewidth} p{0.18\linewidth}}
\toprule
\textbf{Category} & \textbf{Matching rule on command text} & \textbf{Example} \\
\midrule
\texttt{test\_execution} &
  Contains \texttt{pytest}, \texttt{python -m pytest},
  \texttt{make test}, \texttt{cargo test}, or \texttt{npm test}. &
  \texttt{pytest tests/test\_io.py} \\[2pt]
\texttt{file\_edit} &
  Contains \texttt{sed -i}, \texttt{cat >}, \texttt{cat <<},
  \texttt{tee}, \texttt{echo >}, or \texttt{patch}. &
  \texttt{sed -i 's/foo/bar/' f.py} \\[2pt]
\texttt{dependency\_setup} &
  Contains \texttt{pip install}, \texttt{apt-get install},
  \texttt{cmake}, \texttt{gcc}, \texttt{make}, or \texttt{git clone}. &
  \texttt{pip install numpy} \\[2pt]
\texttt{git\_operation} &
  Starts with \texttt{git diff}, \texttt{git log},
  \texttt{git status}, \texttt{git commit}, \texttt{git add},
  \texttt{git show}, or \texttt{git blame}. Excludes \texttt{git clone}
  (classified as dependency setup). &
  \texttt{git diff HEAD\textasciitilde 1} \\[2pt]
\texttt{search} &
  Contains \texttt{grep}, \texttt{rg}, \texttt{ag}, \texttt{ack}, or
  \texttt{find} with \texttt{-name}/\texttt{-type}. &
  \texttt{grep -rn TODO src/} \\[2pt]
\texttt{file\_read} &
  Reads file contents without modifying them: \texttt{cat} (no
  redirect), \texttt{head}, \texttt{tail}, \texttt{sed -n},
  \texttt{less}, or \texttt{more}. &
  \texttt{cat README.md} \\[2pt]
\texttt{script\_execution} &
  Invokes an interpreter: \texttt{python3}, \texttt{python},
  \texttt{node}, \texttt{bash}, \texttt{rscript}, or \texttt{sqlite3}.
  Test patterns are routed to \texttt{test\_execution} first. &
  \texttt{python repro.py} \\[2pt]
\texttt{navigation} &
  Starts with \texttt{ls}, \texttt{pwd}, \texttt{tree}, or \texttt{cd}. &
  \texttt{ls -la src/} \\[2pt]
\texttt{interrupt\_abort} &
  Contains \texttt{C-c}, \texttt{kill}, \texttt{pkill}, or is exactly
  \texttt{q}/\texttt{exit}. &
  \texttt{C-c} (cancel REPL) \\[2pt]
\texttt{reasoning\_only} &
  Action JSON contains no parseable command, or is empty/raw text. &
  Agent ``thinking aloud'' \\
\bottomrule
\end{tabular}
\caption{Action category classification rules. Each turn matches the
highest-priority rule for which any of its commands satisfies the
pattern. \texttt{reasoning\_only} is the residual class for turns
whose action object contains no parseable command.}
\label{tab:category-rules}
\end{table*}

\section{Details of Data Curation}
\label{sec:appen_curation}

Although we use a shared policy to learn control in both directions, the data construction process needs to be designed separately for the two components. Following the benchmark-driven harness optimization protocol of Meta-Harness~\citep{lee2026meta}, we construct supervision from solved SWE-bench Verified trajectories. The supervision is used only for the harness-level projection modules, while the generator model remains frozen. We use Qwen3.5-35B-A3B as the generator and initial harness backbone to produce interaction traces under the HarnessBridge pipeline.

We automate the data construction pipeline by: (1) extensively prompting existing models to generate candidate supervision; and (2) applying strict data-filtering criteria to retain high-quality examples. Our data curation is designed to construct supervision that is both correct and useful for learning harness-level decisions, rather than merely collecting large numbers of raw intervention traces. Starting from approximately 40K raw supervision candidates, we apply separate curation principles for action projection and observation projection.

\textbf{For action projection}, each example specifies whether a proposed action should be passed to the environment or rejected with feedback. We pay particular attention to rejected-action examples, since they are naturally sparse and directly teach the harness when not to expose a proposed action to the environment. Rejection examples are derived from logged rejected actions and include the rejection category, concern, grounding evidence, and suggested revision. We retain rejection examples only when the rejection is trajectory-grounded, followed by a successful subsequent correction, and improves interaction efficiency. In other words, a rejected action is used as supervision only if the rejection is supported by the existing trajectory, the revised behavior remains correct, and the intervention helps avoid unnecessary environment steps or token consumption. This prevents the model from learning arbitrary or unverifiable rejections. Pass examples are constructed from actions that were successfully executed in solved trajectories. To avoid an always-pass bias, we balance the final action-projection set to a 50:50 \textsc{Pass}/\textsc{Reject} ratio, yielding $2{,}682$ examples. We also avoid constructing artificial negative examples by simply flipping previous rejection labels into \textsc{Pass} labels, since we observed that such label flipping creates a substantial distribution shift and encourages over-rejection behavior.

\textbf{For observation projection}, each example consists of a trajectory history and a structured target indicating which turns should be preserved, summarized, or omitted. The target also includes an active-state index that records decision-relevant facts to keep visible to the generator. One of the main objectives is to learn an appropriate compression policy rather than an overly aggressive summarizer. We therefore collect examples under multiple reserved-history windows, ranging from small windows such as $N=10$ to larger windows such as $N=100$, to expose the model to different context-compression regimes. Very small windows tend to force overly harsh summarization, while very large windows provide limited compression signal. We consequently emphasize intermediate regimes, such as $N=30$ and $N=50$, which better capture the desired trade-off between preserving decision-critical context and reducing token cost. We further subsample keep-all cases and balance examples across different compression regimes, so that the model does not collapse into either always preserving the full trajectory or compressing too aggressively. After filtering and deduplication, the final observation-projection set contains $2{,}723$ examples.

After LLM-as-judge filtering with GLM-4.7-Flash, deduplication, trajectory-level capping, rejection-quality validation, and distribution balancing, we retain $5{,}405$ high-quality examples for supervised fine-tuning. Both components are converted into a unified instruction-following format and are used to fine-tune a Qwen3.5-0.8B model with supervised fine-tuning.

\section{Experiment Implementation}
\subsection{More Experiment Setup}
\label{sec:appen_setup}

All experiments are conducted using the Harbor framework, with open-source models served on NVIDIA H200 GPUs.

\paragraph{Meta-Harness.}
We tune the Terminus~2 harness code on 100 tasks sampled from SWE-bench Verified and evaluate on the full SWE-bench Verified and Terminal-Bench~2.0 suites. Tuning is performed for 3 iterations using Qwen-3.5-35B-A3B as the optimization model.

\paragraph{\textsc{HarnessBridge}.}
We set the observation projection window size per backbone to approximately the median trajectory length of the baseline (Terminus~2) on each evaluation benchmark, without reference to \textsc{HarnessBridge}'s own performance. This calibrates the window to each model's interaction profile, accounting for substantial differences in trajectory length. For instance, Claude-Opus-4.7 tends to complete tasks in fewer turns, whereas DeepSeek-V4-Pro variants typically produce longer trajectories. By default, we use a window size of 20, except for Claude-Opus-4.6 (10) and DeepSeek-V4-Pro and DeepSeek-V4-Flash (30); on SWE-bench Verified with Qwen-3.5-35B-A3B, we use 50. The action projection cap is set to 5 per task on Terminal-Bench~2.0 and 10 on SWE-bench Verified.

\paragraph{SFT Training.}
We fine-tune Qwen-3.5-0.8B using the Adam optimizer with a learning rate of $1\mathrm{e}{-5}$, batch size of 64, and bf16 precision for 3 epochs on NVIDIA H200 GPUs.

\paragraph{HarnessBridge Deployment Cost.}
In addition to generator token consumption, we measure the inference overhead introduced by the HarnessBridge controller. HarnessBridge uses a lightweight Qwen3.5-0.8B model for observation and action projection, whereas the evaluated generators are substantially larger open-source or commercial models. Although the controller processes approximately \(3\times\) as many tokens as the generator in our runs, its per-token inference cost is much lower. Under a parameter-normalized compute proxy, the controller costs only \(0.8/35 \approx 2.3\%\) of a 35B dense generator per token. Therefore, even with \(3\times\) more controller tokens, its compute-weighted overhead is approximately \(3 \times 0.8/35 \approx 6.9\%\) of the corresponding 35B-generator cost.

This proxy is conservative for practical deployment: Qwen3.5-0.8B can be served with a low memory footprint and high throughput on commodity accelerators, and its marginal inference cost is small even compared with the open-source Qwen3.5-35B-A3B generator. When paired with commercial API-based generators, the controller is deployed locally and its cost is further amortized relative to the API-side generator cost, making the additional harness computation small in typical end-to-end deployments.

After including all controller input/output tokens, HarnessBridge still reduces total compute-weighted inference cost across the evaluated settings. This indicates that the reported efficiency gain is not merely a transfer of computation from the generator to the harness, but an end-to-end reduction in interaction cost. In practice, the lightweight controller acts as a low-cost interface policy that prevents expensive generator-context growth and unproductive environment interactions, so a modest amount of local harness computation yields a substantially larger reduction in downstream generator inference cost.

\subsection{Baseline Tuning and Evaluation Fairness}
To ensure a fair comparison, all harnesses are evaluated under the same generator backbone, benchmark split, decoding configuration, and task-level execution budget. 
For each generator, we use the same serving backend and temperature setting across harnesses, and we keep the benchmark tasks, timeout constraints, and success criteria fixed. We do not tune any harness on the evaluation trajectories. For baselines that expose no learning or search procedure, including Terminus 2, mini-SWE-agent, OpenHands, and Qwen-Coder, we use their official or recommended configurations and only adapt interface-level details necessary to connect them to the same generator and benchmark environment. We do not perform additional benchmark-specific prompt or policy optimization for these fixed baselines.

This protocol is intended to separate harness-level optimization from test-time evaluation: each method is allowed to use its standard harness construction mechanism, but all methods are compared under the same frozen generators, same task suites, and same execution constraints.

\subsection{Data Curation for \textsc{HarnessBridge-D}}
\label{sec:appen_curation_d}

For \textsc{HarnessBridge-D}, we follow the curation principles of
Appendix~\ref{sec:appen_curation}, with adaptations motivated by the
characteristics of DeepSeek-V4 trajectories. As noted in
Section~\ref{sec:appen_setup}, DeepSeek-V4 variants produce substantially longer and
more reasoning-heavy trajectories than Qwen3.5-35B-A3B, which makes both
the verbosity of individual turns and the length of the trajectory history
the dominant sources of low-quality supervision. As before, we retain
supervision only from solved trajectories, so that every example is
grounded in a correct task completion.

\textbf{Action projection.} We draw action-projection supervision only from
trajectories in which \textsc{HarnessBridge-D} solves the task, and we
condition the retention criterion on the baseline (Terminus~2) outcome. For
\emph{both-pass} tasks, where the baseline already solves the task, the
controller contributes only efficiency; we therefore retain such
trajectories as supervision only when they reduce both token usage and
interaction turns relative to the baseline. For \emph{fail-to-pass} tasks, where the
baseline fails but \textsc{HarnessBridge-D} succeeds, the trajectory already
demonstrates that harness control enables a new solve; we apply a looser
criterion and retain it when it saves either tokens or turns. In both cases
we exclude trajectories that yield no efficiency gain, since training on
turn-increasing examples teaches the validator to intervene where
intervention is unwarranted, producing over-rejection.

We validate this criterion in Table~\ref{tab:curation_ablation}. The
\emph{turn-agnostic} recipe retains all correct trajectories, including
turn-increasing ones, and balances the set 50:50 (reject:pass); the
\emph{turn-saving} recipe retains only efficiency-improving trajectories,
which shifts the natural balance to 7:10. The turn-agnostic recipe degrades
success rate \emph{below} the no-harness baseline on both DeepSeek-V4-Pro
($57.3 \rightarrow 48.3$) and DeepSeek-V4-Flash ($49.4 \rightarrow 43.8$),
with a high validator rejection rate ($\sim$37\%) and frequent timeouts. The
turn-saving recipe instead improves over the baseline ($57.3 \rightarrow
59.6$ and $49.4 \rightarrow 51.7$) while roughly halving the rejection rate
and cutting timeouts. A validator trained on turn-increasing supervision
over-rejects to the point of making the harness worse than the baseline Terminus 2;
curating toward genuinely efficiency-improving trajectories is therefore
essential. The turn-saving recipe is the one we report as
\textsc{HarnessBridge-D}.

\textbf{Observation projection.} Each example again consists of a
trajectory history and a structured target marking which turns to preserve,
summarize, or drop. Given the verbosity of DeepSeek-V4 trajectories, we add
two length-oriented filters. First, we discard any example whose target
output exceeds 6K tokens, since such targets correspond to summaries that
are themselves too long to provide a useful compression signal. Second, we
require that every structured target parse successfully into the expected
schema, discarding malformed targets. To prevent a small number of very
long trajectories from dominating the set, we cap supervision at 20
examples per trajectory. Finally, to preserve decision-critical recent
context, we protect the most recent turns from compression: the last fewer
than three turns are never dropped, and no summarization entry is applied
within that recent window. After filtering and deduplication, the final
observation-projection set contains 3575 examples.

After LLM-as-judge filtering with DeepSeek-V4-Pro, deduplication,
trajectory-level capping, and the length- and efficiency-based criteria
above, we retain 5287 high-quality examples for fine-tuning the
\textsc{HarnessBridge-D} controller.

\begin{table}[t]
\centering
\begin{tabular}{llcc}
\toprule
Generator & Recipe & SR (\%) & Reject rate (\%) \\
\midrule
\multirow{3}{*}{DeepSeek-V4-Pro}
  & Terminus 2    & 57.3 & -- \\
  & Turn-agnostic & 48.3 & 36.7 \\
  & Turn-saving   & \textbf{59.6} & 18.3 \\
\midrule
\multirow{3}{*}{DeepSeek-V4-Flash}
  & Terminus 2    & 49.4 & -- \\
  & Turn-agnostic & 43.8 & 37.2 \\
  & Turn-saving   & \textbf{51.7} & 19.3 \\
\bottomrule
\end{tabular}
\caption{Effect of action-projection curation recipe on Terminal-Bench~2.0.
\emph{Turn-agnostic} includes correct trajectories that increase turns
(50:50 reject:pass); \emph{turn-saving} retains only turn-saving
trajectories (7:10 reject:pass). The turn-agnostic recipe falls below the
no-harness baseline; the turn-saving recipe is reported as
\textsc{HarnessBridge-D}.}
\label{tab:curation_ablation}
\end{table}

\subsection{Experiment Result for \textsc{HarnessBridge-D}}
\label{sec:appen_result_d}
\begin{table*}[t]
\centering
\small
\setlength{\tabcolsep}{1.7pt}
\renewcommand{\arraystretch}{1.12}
\begin{tabular}{lcc cc cc cc}
\toprule
\multirow{2}{*}{\textbf{Harness}}
& \multicolumn{2}{c}{\textbf{GPT-5.4-Nano}}
& \multicolumn{2}{c}{\textbf{GPT-5.4}}
& \multicolumn{2}{c}{\textbf{DeepSeek-V4-Flash}}
& \multicolumn{2}{c}{\textbf{DeepSeek-V4-Pro}} \\
\cmidrule(lr){2-3}
\cmidrule(lr){4-5}
\cmidrule(lr){6-7}
\cmidrule(lr){8-9}
& SR $\uparrow$& Token $\downarrow$
& SR $\uparrow$& Token $\downarrow$
& SR $\uparrow$& Token $\downarrow$
& SR $\uparrow$& Token $\downarrow$ \\
\midrule
Terminus~2
& 18.0 & 9.80
& 53.9 & 9.41
& 49.4 & 2.18
& 57.3 & 1.02 \\
\textsc{HarnessBridge-D}
& 23.6\pos{31.1\%} & 1.00\post{89.8\%}
& 52.8\nega{2.0\%} & 0.51\post{94.6\%}
& 51.7\pos{4.7\%} & 0.92\post{57.8\%}
& 59.6\pos{4.0\%} & 0.98\post{3.9\%} \\
\bottomrule
\end{tabular}
\caption{Success rate (SR) and average input token usage (in millions) on Terminal-Bench~2.0 across models using \textsc{HarnessBridge-D}.}
\label{tab:agent_harness_results}
\vspace{-0.5cm}
\end{table*}
Table~\ref{tab:agent_harness_results} reports success rate (SR) and average
input token usage on Terminal-Bench~2.0. Trained only on DeepSeek-sampled
supervision, \textsc{HarnessBridge-D} substantially reduces input token
consumption across all generators while maintaining comparable SR. The
reductions are largest on the GPT-5.4 family, whose unconstrained baselines
are the most token-heavy ($9.80$M and $9.41$M input tokens): on GPT-5.4-Nano
it cuts usage by roughly $90\%$ (to $1.00$M), with a similar reduction on
GPT-5.4. On DeepSeek-V4-Pro, whose baseline is already token-lean
($1.02$M), the reduction is naturally smaller ($-3.9\%$). The largest savings appear on the models with the longest baseline
trajectories (the GPT-5.4 family). More broadly, these results suggest that
the effectiveness of HarnessBridge is largely independent of the
data-sampling model: harness-control supervision drawn from different
generators yields a similarly strong, transferable controller.

\section{Limitation}
Our evaluation is confined to coding settings. The harness-control mechanism
is not specific to code, however: observation and action projection operate
over generic tool-use trajectories and manage long interaction histories
rather than any code-specific structure, so we expect the approach to extend
to other tool-heavy, long-horizon agentic domains such as web navigation,
computer use, and multi-step research workflows. Due to evaluation cost, we report single-run results and
accordingly emphasize relative trends over absolute values. Empirically validating this
generalization remains an important direction for future work.
\providecolor{promptbg}{RGB}{235,243,252}      
\providecolor{promptborder}{RGB}{45,95,165}    
\providecolor{prompttitle}{RGB}{20,50,110}     
\providecolor{codebg}{RGB}{220,232,246}        
\providecolor{codeborder}{RGB}{130,165,210}    
\providecolor{stringcolor}{RGB}{30,80,160}     

\lstdefinelanguage{jsonPrompt}{
  basicstyle=\ttfamily\small,
  numbers=none,
  showstringspaces=false,
  breaklines=true,
  frame=single,
  backgroundcolor=\color{codebg},
  rulecolor=\color{codeborder},
  string=[s]{"}{"},
  stringstyle=\color{stringcolor},
  comment=[l]{//},
  commentstyle=\color{black!60},
}

\makeatletter
\@ifundefined{promptheading}{%
  \newcommand{\promptheading}[1]{%
    \vspace{6pt}{\large\bfseries\color{prompttitle} #1}\par\vspace{2pt}}%
}{}
\@ifundefined{partheading}{%
  \newcommand{\partheading}[1]{%
    \vspace{3pt}{\small\bfseries\color{prompttitle} #1}\par\vspace{1pt}}%
}{}
\@ifundefined{axisheading}{%
  \newcommand{\axisheading}[1]{%
    \vspace{4pt}{\normalsize\bfseries\color{prompttitle} #1}\par\vspace{1pt}}%
}{}

\tcbset{promptbox@style/.style={%
    enhanced, breakable,
    colback=promptbg, colframe=promptborder,
    colbacktitle=promptborder, coltitle=white,
    fonttitle=\bfseries\large,
    attach boxed title to top left={xshift=10pt, yshift=-10pt},
    boxed title style={sharp corners, colframe=promptborder, colback=promptborder},
    boxrule=1pt, arc=4pt, left=14pt, right=14pt, top=18pt, bottom=12pt}}
\@ifundefined{promptbox}%
  {\newtcolorbox{promptbox}[1][]{promptbox@style, title={#1}}}%
  {\renewtcolorbox{promptbox}[1][]{promptbox@style, title={#1}}}
\makeatother


\section{Instruction Prompt}

\begin{promptbox}[PROMPT (Observation Projection)]

\promptheading{Role}

Your job is to manage the conversation history so the agent stays focused and within the token budget. Remove or summarize redundant and irrelevant tool responses and conversation. Preserve precise and important information.

\promptheading{Input (JSON)}

\begin{itemize}[leftmargin=*, itemsep=2pt]
  \item \textbf{task}: original task instruction
  \item \textbf{turn}: current turn number (1-indexed)
  \item \textbf{history}: full list of prior turns, each containing:
    \begin{itemize}[leftmargin=*, itemsep=1pt]
      \item \texttt{turn}: turn number
      \item \texttt{reasoning}: the agent's THOUGHT/analysis before acting
      \item \texttt{action}: the command the agent executed
      \item \texttt{observation}: the terminal output from that command
      \item \texttt{tokens}: token count for this turn
    \end{itemize}
  \item \texttt{history} contains all completed turns. \texttt{last\_observation} is NOT in history --- it is the new observation from the current turn, not yet appended.
  \item \textbf{last\_observation}: terminal output from the current turn's command, not yet in history
  \item \textbf{total\_tokens}: current conversation token count
  \item \textbf{budget}: soft token target
\end{itemize}

\promptheading{Your Job}

\begin{enumerate}[leftmargin=*, itemsep=4pt]
  \item Decide which turns to \textbf{SUMMARIZE} or \textbf{DROP}. Any turn you do not mention is kept unchanged --- do not list turns you are keeping.

  \item Extract an \textbf{active-state index} from prior turns. Include any decision-relevant items that should remain visible, such as unresolved errors, open tests, confirmed facts, constraints, pending subtasks, important files/functions, hypotheses, prior decisions, failed approaches to avoid, or partial progress. This index may contain items from multiple turns and is placed before the history.
\end{enumerate}

The harness assembles the final prompt: active-state index at the top, then history in chronological order (keeping originals by default, replacing summarized turns with your summaries, and skipping dropped turns). \texttt{last\_observation} is appended separately by the harness. The harness handles formatting --- write summaries as plain text.

\promptheading{Principles}

\begin{itemize}[leftmargin=*, itemsep=4pt]
  \item \textbf{Always keep}: system prompt, task description, and the most recent turns (the harness protects these automatically). Do not summarize or drop any recent turns.

  \item \textbf{Bias toward keeping} turns containing: test output, compilation errors, numerical results, error messages with codes/paths/line numbers, strategy changes, code snippets, function signatures. These lose critical detail when summarized.

  \item \textbf{Summarize or drop older turns} only when they have been clearly superseded, resolved, or duplicated by later turns.

  \item \textbf{When summarizing}:
    \begin{itemize}[leftmargin=*, itemsep=1pt]
      \item Preserve exact values verbatim: file paths, line numbers, error codes, function names, variable names. Never paraphrase these.
      \item Include what was tried, what resulted, and whether it advanced the task.
      \item 1--3 sentences. Shorter is better if the turn was minor.
    \end{itemize}
\end{itemize}

\promptheading{Output}

Valid JSON, nothing else. Only list turns you are changing. Turns not listed are kept unchanged.

\begin{lstlisting}[language=jsonPrompt]
{
  "summarize": [
    {"turn": N, "content": "1-3 sentence summary with exact values"},
    ...
  ],
  "drop": [
    {"turn": N, "reason": "1-3 sentence reason for dropping this turn"},
    ...
  ],
  "active_state": [
    {"turn": N, "content": "one-line decision-relevant item with exact values"}
  ]
}
\end{lstlisting}

\end{promptbox}

\begin{promptbox}[PROMPT (Action Projection)]

\promptheading{Role}

You are a decision verifier for an AI coding agent. Before the agent's proposed command executes, you assess whether it is a sound use of the agent's limited turns and budget given the task and trajectory so far.

\promptheading{Input (JSON)}

\begin{itemize}[leftmargin=*, itemsep=2pt]
  \item \textbf{task}: original task instruction
  \item \textbf{history}: full list of prior turns, each containing:
    \begin{itemize}[leftmargin=*, itemsep=1pt]
      \item \texttt{turn}: turn number
      \item \texttt{action}: the command the agent executed (may embed the agent's prior reasoning as JSON)
      \item \texttt{observation}: the terminal output from that command
    \end{itemize}
  \item \textbf{agent\_reasoning}: the agent's analysis/plan for the proposed command
  \item \textbf{proposed\_command}: the command about to be executed
  \item \textbf{is\_submission}: whether this is a submission/task-complete action
\end{itemize}

\promptheading{Decide: PASS or REJECT}

REJECT when the proposed command is a clear misuse of the agent's turns. Common patterns include (non-exhaustive --- other concerns qualify if you can meet the same evidence bar):

\begin{itemize}[leftmargin=*, itemsep=6pt]
  \item \textbf{SHALLOW\_EDIT}: The agent is about to make an edit that does not engage with the actual mechanism the task describes --- pattern-matching on surface keywords or touching adjacent code rather than tracing the real logic path the task requires.

  \item \textbf{PREMATURE\_SUBMISSION}: The agent is submitting a fix without having verified that it addresses the core issue described in the task, or the agent's own observations suggest the fix is incomplete or wrong.

  \item \textbf{WASTEFUL\_EXPLORATION}: The agent has already identified the relevant code location and mechanism (evidenced by prior edits, specific file/line references in reasoning, or explicit statements about the bug location), but is spending 3+ additional turns on further reading or searching that does not target the identified location.

  \item \textbf{CONTRADICTED\_PATH}: The agent's own earlier observations contain evidence (error messages, test failures, incorrect output) that the current approach is flawed, and the agent has not addressed or acknowledged that evidence in its reasoning --- yet is continuing the same logical approach with a different command.
\end{itemize}

\promptheading{Mitigating Factors}

Before rejecting, consider whether any of these mitigating factors apply:

\begin{itemize}[leftmargin=*, itemsep=4pt]
  \item The command targets the logical path the task describes, even if you would implement it differently.
  \item The agent's reasoning demonstrates it traced the relevant mechanism, even briefly.
  \item The task is straightforward enough that direct action without deep tracing is appropriate.
  \item The agent has not yet identified a specific bug location or mechanism and is still in legitimate exploration.
  \item This is the agent's first attempt at an edit or fix --- allow it to try and learn from the result rather than pre-empting it.
  \item The projection has already rejected the same category of issue 3+ times in this trajectory and the agent has not changed approach. Further rejections of the same pattern are unlikely to help --- PASS and let the agent attempt execution so it can learn from actual failure output.
  \item You cannot articulate a specific concern backed by evidence from the task or the history.
\end{itemize}

\textbf{Default to PASS. When in doubt, PASS.} The cost of a false reject (wasted turn + lost momentum) is higher than the cost of letting a questionable command through (the agent can course-correct from a bad result, but cannot recover a rejected turn).

On REJECT you MUST populate \texttt{concern}, \texttt{evidence}, and \texttt{suggestion} with specific references. The \texttt{evidence} field must include a direct quote from the task or history that contradicts the proposed command. If you cannot provide a direct quote, you must PASS.

\promptheading{Output}

Valid JSON, nothing else.

On PASS, emit only the decision:

\begin{lstlisting}[language=jsonPrompt]
{"decision": "PASS"}
\end{lstlisting}

On REJECT, populate every field:

\begin{lstlisting}[language=jsonPrompt]
{
  "decision": "REJECT",
  "category": "SHALLOW_EDIT" |"PREMATURE_SUBMISSION"|"WASTEFUL_EXPLORATION"| "CONTRADICTED_PATH",
  "concern": "what the command fails to address or why it is wasteful",
  "evidence": "specific quote from task, observation, or agent reasoning that supports the concern",
  "suggestion": "concrete next step - a specific file to read, function to trace, command to run, or assumption to verify"
}
\end{lstlisting}

\end{promptbox}

\begin{promptbox}[LLM Judge PROMPT (Action Projection)]

\promptheading{Role}

You are an expert evaluator scoring training examples for an LLM action-projection model. The model's job is to project whether a proposed agent command should PROCEED or BE BLOCKED before it executes. You will be shown the official projection policy (tolerant or strict), the model's INPUT (\texttt{task} + \texttt{history} + \texttt{agent\_reasoning} + \texttt{proposed\_command} + \texttt{is\_submission}), and the model's OUTPUT (\texttt{decision} + \texttt{category} + \texttt{concern} + \texttt{evidence} + \texttt{suggestion}).

Your job: score this (input, output) pair on seven axes from 1 to 5, and return STRICT JSON. Use the rubrics below.

\promptheading{Policy Reference}

\texttt{\{policy\}}

\promptheading{Axis Rubrics}

\axisheading{1. INPUT WORTHINESS (1-5) -- gates the record}

Decide whether this INPUT is one the model should be trained on at all --- i.e. whether the \texttt{proposed\_command} is genuinely a decision-worthy moment. Look ONLY at the input here.

\begin{itemize}[leftmargin=*, itemsep=2pt]
  \item \textbf{5}: Decision is meaningfully borderline. Multiple of:
    \begin{itemize}[leftmargin=*, itemsep=1pt]
      \item \texttt{proposed\_command} is a real edit (\texttt{sed -i}, write, patch), a build/install/run command with side effects, or a submission (\texttt{is\_submission=true})
      \item History contains substantive context: prior edits, observed test results, error messages, or the agent has made claims about the bug location
      \item \texttt{agent\_reasoning} is non-trivial and shows a plan (not empty, not ``I will run the command'')
      \item Both PROCEED and BLOCK are defensible after careful reading, OR the case is a clear BLOCK for a non-obvious reason
    \end{itemize}
  \item \textbf{4}: Genuine edit/submission with adequate context, but the decision is fairly clear (one side is much stronger).
  \item \textbf{3}: Borderline. Edit is real but the call is obvious, OR the history is thin but the command is substantive enough to teach something.
  \item \textbf{2}: Marginal training value: edit is small/cosmetic, OR history is too thin to ground a real decision, OR \texttt{agent\_reasoning} is empty/boilerplate.
  \item \textbf{1}: Trivial input. Examples that score 1:
    \begin{itemize}[leftmargin=*, itemsep=1pt]
      \item \texttt{proposed\_command} is a whitelisted simple read (\texttt{cat} / \texttt{ls} / \texttt{grep} / \texttt{pwd}) --- L2 should not have fired anyway
      \item First turn of the trajectory with no prior context, on a read-only command
      \item \texttt{agent\_reasoning} is empty AND history is \textless{} 3 turns
      \item The command is so obviously fine that any model would PROCEED (e.g. running the test harness exactly as the task asks)
    \end{itemize}
\end{itemize}

\textbf{GUIDANCE}: 25\% of records are PROCEED with synthetic null bodies. Many of these are on uninteresting commands. Use this axis to demote them --- even a perfectly-decided trivial PROCEED is poor SFT signal.

\axisheading{2. SCHEMA (1--5)}

\begin{itemize}[leftmargin=*, itemsep=2pt]
  \item \textbf{5}: Valid JSON with exactly five keys: \texttt{decision}, \texttt{category}, \texttt{concern}, \texttt{evidence}, \texttt{suggestion}. \texttt{decision} $\in$ \{\texttt{"PROCEED"}, \texttt{"BLOCK"}\}. On PROCEED, the other four are null. On BLOCK, \texttt{concern} + \texttt{evidence} + \texttt{suggestion} are non-null strings.
  \item \textbf{3}: Parses but one optional field misshapen.
  \item \textbf{1}: Does not parse, OR BLOCK with empty \texttt{concern}/\texttt{evidence}/\texttt{suggestion}, OR PROCEED with non-null fields.
\end{itemize}

\axisheading{3. DECISION\_CORRECTNESS (1--5)}

Independently judge whether the proposed command should PROCEED or BLOCK given the policy, the task, and the history.

\begin{itemize}[leftmargin=*, itemsep=2pt]
  \item \textbf{5}: The output's decision matches your independent judgment.
  \item \textbf{3}: Borderline case --- either decision defensible.
  \item \textbf{1}: Clear disagreement (e.g. output BLOCKs a first-attempt edit that the policy says to allow, OR PROCEEDs on a submission with observed test failures).
\end{itemize}

\textbf{ANCHOR}: When \texttt{agent\_reasoning} explicitly traces the relevant file and line and the command targets that location, PROCEED is correct. When the agent has been blocked on the same category 3+ times already, PROCEED is correct (per the mitigating-factors clause).

\axisheading{4. EVIDENCE\_GROUNDEDNESS (1--5) --- most important quality axis on BLOCKs}

Only scored when \texttt{decision == BLOCK}. The policy mandates a direct quote from the task or history. Verify the \texttt{evidence} string actually appears (or near-appears, allowing for small reformatting) in the provided input.

\begin{itemize}[leftmargin=*, itemsep=2pt]
  \item \textbf{5}: Evidence contains a verbatim quote from \texttt{input.history}, \texttt{input.task}, or \texttt{input.agent\_reasoning} that genuinely supports the concern.
  \item \textbf{3}: Evidence paraphrases the input but the underlying claim is true.
  \item \textbf{2}: Evidence is generic (``the agent has explored too much'') without a specific quote.
  \item \textbf{1}: Evidence references something that does NOT appear in the input --- fabricated quote.
\end{itemize}

On PROCEED records, score this axis 5 (N/A).

\axisheading{5. CONCERN\_SPECIFICITY (1--5)}

Only scored on BLOCK.

\begin{itemize}[leftmargin=*, itemsep=2pt]
  \item \textbf{5}: \texttt{concern} names a specific file, function, line, or behavior and explains \emph{why} the command is unsound.
  \item \textbf{3}: Names the issue at policy-category level but no specifics.
  \item \textbf{1}: Generic platitude (``this is wasteful'').
\end{itemize}

N/A on PROCEED $\rightarrow$ score 5.

\axisheading{6. SUGGESTION\_ACTIONABILITY (1--5)}

Only scored on BLOCK.

\begin{itemize}[leftmargin=*, itemsep=2pt]
  \item \textbf{5}: \texttt{suggestion} is a concrete next step the agent can execute --- a specific file to read, function to trace, command to run, or assumption to verify.
  \item \textbf{3}: Direction is correct but underspecified (``look at the models layer'').
  \item \textbf{1}: Empty, or just restates the concern.
\end{itemize}

N/A on PROCEED $\rightarrow$ score 5.

\axisheading{7. CATEGORY\_FIT (1--5)}

Only scored on BLOCK.

\begin{itemize}[leftmargin=*, itemsep=2pt]
  \item \textbf{5}: \texttt{category} matches the \texttt{concern}. \texttt{WASTEFUL\_EXPLORATION} is used only when prior reasoning identified a location and the proposed command does not target it. \texttt{SHALLOW\_EDIT} is used only when the edit does not engage the mechanism the task names. \texttt{PREMATURE\_SUBMISSION} only on submissions. \texttt{CONTRADICTED\_PATH} only when prior observations contradict the current approach.
  \item \textbf{3}: Category is plausible but a different category fits better.
  \item \textbf{1}: Category contradicts the concern (e.g. \texttt{WASTEFUL\_EXPLORATION} on a first-turn search).
\end{itemize}

N/A on PROCEED or \texttt{category=null} $\rightarrow$ score 5.

\axisheading{8. BIAS\_SAFETY (1--5)}

Detect failure modes the SFT corpus is known to over-represent.

\begin{itemize}[leftmargin=*, itemsep=2pt]
  \item \textbf{5}: This record does NOT exhibit any of the biases below.
  \item \textbf{3}: One mild bias signal.
  \item \textbf{1}: Strong bias signal, e.g.:
    \begin{itemize}[leftmargin=*, itemsep=1pt]
      \item BLOCK \texttt{WASTEFUL\_EXPLORATION} on the first 3 turns of a trajectory before the agent has had a fair chance to explore
      \item BLOCK on a first-attempt edit (policy says allow first attempts)
      \item PROCEED with a synthesized null body when the trajectory actually contained reasoning the model could have articulated
      \item BLOCK category that does not match what the concern describes (overfitting to category labels)
    \end{itemize}
\end{itemize}

\promptheading{Overall Score (BLOCK records)}

\begin{lstlisting}[language=jsonPrompt, basicstyle=\ttfamily\small]
overall = round( (worthiness*2 + schema*1 + correctness*2 + grounded*2
                + concern*1 + suggestion*1 + cat_fit*1 + bias*1.5)
                / 11.5, 2 )
\end{lstlisting}

\promptheading{Overall Score (PROCEED records)}

\begin{lstlisting}[language=jsonPrompt, basicstyle=\ttfamily\small]
overall = round( (worthiness*2 + schema*1 + correctness*3 + bias*1.5)
                / 7.5, 2 )
\end{lstlisting}

(\texttt{grounded} / \texttt{concern} / \texttt{suggestion} / \texttt{cat\_fit} are N/A and not weighted.)

\promptheading{Keep Decision}

For BLOCK:

\begin{lstlisting}[language=jsonPrompt, basicstyle=\ttfamily\small]
keep = (input_worthiness >= 3) AND
       (overall >= 3.5) AND
       (grounded >= 4) AND
       (correctness >= 4)
\end{lstlisting}

For PROCEED:

\begin{lstlisting}[language=jsonPrompt, basicstyle=\ttfamily\small]
keep = (input_worthiness >= 3) AND
       (overall >= 3.5) AND
       (correctness >= 4) AND
       (the output is NOT the synthetic empty
        {"decision":"PROCEED","category":null,"concern":null,
         "evidence":null,"suggestion":null} when the
        history contains substantive reasoning)
\end{lstlisting}

If the PROCEED body is the synthetic empty form, set \texttt{keep = false} and add flag \texttt{"synthetic\_proceed\_no\_reasoning"}.

\textbf{Note}: an \texttt{input\_worthiness == 1} record (e.g. \texttt{proposed\_command} is \texttt{cat foo.py} with no history) is dropped \emph{even if the decision is correct and the schema is perfect}. The model cannot generalize from trivial decisions.

\promptheading{Output Format}

Emit only this JSON, nothing else:

\begin{lstlisting}[language=jsonPrompt]
{
  "input_worthiness": <int 1-5>,
  "schema":  <int 1-5>,
  "decision_correctness": <int 1-5>,
  "evidence_groundedness": <int 1-5>,
  "concern_specificity": <int 1-5>,
  "suggestion_actionability":<int 1-5>,
  "category_fit": <int 1-5>,
  "bias_safety":  <int 1-5>,
  "overall": <float>,
  "keep_for_sft": <bool>,
  "rationale": "<2-4 sentences. Begin with one sentence on whether the proposed_command + history merit a projection decision at all, then cite the quote that grounds (or fails to ground) the evidence>",
  "flags":  ["<short tag>", ...]
    // e.g. trivial_input_simple_read,
    // thin_history_no_context,
    // fabricated_quote,
    // wasteful_first_3_turns,
    // synthetic_proceed_no_reasoning,
    // category_mismatch
}
\end{lstlisting}

\end{promptbox}

\begin{promptbox}[LLM Judge PROMPT (Observation Projection)]

\promptheading{Role}

You are an expert evaluator scoring training examples for an LLM context-curator. The curator compresses an AI coding agent's chat history without losing critical detail. You will be shown the official curator policy, the curator's INPUT (\texttt{task} + raw \texttt{history} + \texttt{last\_observation} + token state), and the curator's OUTPUT (\texttt{summarize} / \texttt{drop} / \texttt{unresolved} / \texttt{curated\_last\_observation}).

Your job: score this (input, output) pair on seven axes from 1 to 5 and return STRICT JSON. The rubrics below center on FIVE specific concerns:

\begin{itemize}[leftmargin=*, itemsep=4pt]
  \item \textbf{C1: PER-TURN DECISION CORRECTNESS.} The curator chose ONE OF THREE actions for every turn in \texttt{input.history}: KEEP (turn is in neither array), SUMMARIZE (turn is in \texttt{output.summarize}), or DROP (turn is in \texttt{output.drop}). For EVERY turn, decide whether the chosen action was right for that turn's content. This is symmetric --- KEEP decisions are scored just like SUMMARIZE and DROP decisions. Passthrough records are simply records where every turn was assigned KEEP; each KEEP is checked.

  \item \textbf{C2: UNRESOLVED INDEX}, both directions.
    \begin{itemize}[leftmargin=*, itemsep=1pt]
      \item[(a)] Each entry in \texttt{output.unresolved} is a real open issue at this turn (not fabricated, not already-resolved, exact values preserved).
      \item[(b)] Every real open issue in \texttt{input.history} is listed in \texttt{output.unresolved}.
    \end{itemize}

  \item \textbf{C3: SUMMARY QUALITY.} For each entry in \texttt{output.summarize}: verbatim values preserved (paths, line numbers, error codes, function names), 1--3 sentences, captures what was tried / what resulted / whether it advanced the task.

  \item \textbf{C4: CROSS-REFERENCE INTEGRITY} (added).
    \begin{itemize}[leftmargin=*, itemsep=1pt]
      \item[(a)] No unresolved entry cites a turn that is in \texttt{output.drop} --- such an entry is broken because the cited turn vanishes.
      \item[(b)] Summaries do not duplicate content already captured in unresolved entries (avoid token-budget waste).
      \item[(c)] Multiple summaries / unresolved entries do not contradict each other on shared facts.
    \end{itemize}

  \item \textbf{C5: BUDGET COMPLIANCE} (added). When \texttt{input.total\_tokens} exceeds \texttt{budget}, the curator's decisions must materially reduce the resulting context size. An over-budget input with \textgreater 80\% KEEP decisions indicates failure to compress.
\end{itemize}

Plus universal floors: schema, \texttt{last\_observation} curation, and recent-turn protection (last \texttt{max(5, R/3)} turns are off-limits).

\promptheading{Policy Reference}

\texttt{\{policy\}}

\promptheading{Axis Rubrics}

\axisheading{1. INPUT\_WORTHINESS (1--5) --- serves Concern 4}

Determines whether this record's INPUT can teach the model anything beyond the policy's hardcoded passthrough rule. The key signal is whether input length puts the curator in a regime where a real decision is required.

Compute these signals from the input:
\begin{itemize}[leftmargin=*, itemsep=1pt]
  \item $H$ = \texttt{len(input.history)}
  \item $R$ = \texttt{metadata.reserved\_turns} (default 15)
  \item $T$ = \texttt{input.total\_tokens}
  \item $B$ = \texttt{input.budget} (default 50000)
\end{itemize}

\begin{itemize}[leftmargin=*, itemsep=2pt]
  \item \textbf{5}: Input is squarely in the decision-required regime. $H > R + 5$ AND $T$ is in $[B \cdot 0.5,\ B \cdot 1.0]$. Either compression or careful passthrough is defensible --- the curator's choice carries information.
  \item \textbf{4}: $H > R$ AND $T$ is in $[B \cdot 0.4,\ B \cdot 1.0]$. Decision matters even if one answer is mildly preferred.
  \item \textbf{3}: $H$ is just above $R$, OR $T$ is in $[B \cdot 0.3,\ B \cdot 0.5]$. A mild decision is required.
  \item \textbf{2}: $H \approx R+1$ to $R+2$ AND $T < B \cdot 0.4$. The curator barely fires; passthrough is the obvious default.
  \item \textbf{1}: $H \leq R$, OR $T < B \cdot 0.2$. The example cannot teach anything non-trivial --- at this size, ANY policy would passthrough.
\end{itemize}

This axis is the LOAD-BEARING gate for Concern 4: a passthrough record whose input scores 1--2 here is dropped because passthrough was the trivial-default answer, not a learned restraint.

\axisheading{2. SCHEMA (1--5)}

\begin{itemize}[leftmargin=*, itemsep=2pt]
  \item \textbf{5}: Single JSON object with exactly four keys (\texttt{summarize}, \texttt{drop}, \texttt{unresolved}, \texttt{curated\_last\_observation}). \texttt{summarize} entries are \texttt{\{turn, content\}}. \texttt{unresolved} entries are \texttt{\{turn, summary\}}.
  \item \textbf{3}: Parses but one minor field deviation.
  \item \textbf{1}: Does not parse, or wrong shape.
\end{itemize}

\axisheading{3. DECISION\_APPROPRIATENESS (1--5) --- serves Concerns 1, 4, 5}

The heart of the curator judgment. This axis treats each input turn as receiving exactly one of three labels:

\begin{itemize}[leftmargin=*, itemsep=1pt]
  \item \textbf{KEEP} \quad (turn appears in NEITHER \texttt{summarize} nor \texttt{drop})
  \item \textbf{SUMMARIZE} \quad (turn appears in \texttt{output.summarize})
  \item \textbf{DROP} \quad (turn appears in \texttt{output.drop})
\end{itemize}

Every turn in \texttt{input.history} is implicitly labeled by the curator's output. The judge verifies the correctness of every label, plus three structural sub-checks.

\partheading{PART 1 --- Determine BEST action for each turn}

For each turn $T$ in \texttt{input.history}, read $T$'s action and observation, THEN read all later turns in \texttt{input.history} (and the \texttt{last\_observation}) to see what the agent actually did with $T$'s information. Determine $T$'s best action:

\textbf{BEST[T] = KEEP} when $T$ carries unique signal that LATER turns reference or rely on:
\begin{itemize}[leftmargin=*, itemsep=1pt]
  \item Error codes, traceback paths, line numbers later traced
  \item Test results (PASS/FAIL with specific assertions later inspected)
  \item Function/class signatures later modified
  \item File path discoveries the agent later edits
  \item Strategy decisions that explain later behavior
  \item Numeric values, measurements later compared against
  \item Any value that appears verbatim in a later observation
\end{itemize}

\textbf{BEST[T] = SUMMARIZE} when $T$ carries SOME high-level value AND is verbose AND its details are now consolidated by later turns:
\begin{itemize}[leftmargin=*, itemsep=1pt]
  \item Long traceback whose fix is shown in a later turn (keep ``fix at \texttt{\textless file\textgreater :\textless line\textgreater} was X, error was \texttt{\textless ExceptionType\textgreater}''; drop the verbatim traceback)
  \item Multi-line \texttt{ls}/\texttt{find}/\texttt{grep} output where the agent later acted on a specific subset (keep the choice + rejected alternatives; drop the full output)
  \item Multi-page test output that resolves to one specific failure (keep the failure summary; drop the full pytest dump)
  \item Strategy-exploration turn where the agent considered multiple approaches and chose one (keep the choice + alternatives)
  \item Cumulative diff / git-status whose net result is described by a later commit/edit
\end{itemize}

\textbf{BEST[T] = DROP} when $T$ carries NO information the agent uses or refers back to:
\begin{itemize}[leftmargin=*, itemsep=1pt]
  \item Banner-only / progress-bar-only turns
  \item Repeated identical observations (3+ greps over same dir)
  \item Verbose listings the agent did NOT subsequently act on
  \item Parse-error retries already corrected by a later turn
  \item Rejected commands whose retries succeed with no dependency on the failed attempt
  \item Duplicate edits/observations
\end{itemize}

The protected recent window --- turns in the last \texttt{max(5, R/3)} positions --- has BEST = KEEP by policy fiat (do not evaluate them on content; their KEEP is mandatory).

\partheading{PART 2 --- Compare CHOSEN to BEST per turn}

Build a confusion-matrix view across all turns. Each cell carries an error severity:

\begin{itemize}[leftmargin=*, itemsep=1pt]
  \item \textbf{correct} --- no penalty.
  \item \textbf{under-comp} --- KEEP when SUMMARIZE was best, OR SUMMARIZE when DROP was best. Bloat but no info loss. Mild penalty.
  \item \textbf{under-comp+} --- KEEP when DROP was best. Significant noise retained. Moderate penalty.
  \item \textbf{over-cost} --- SUMMARIZE when DROP was best. Wastes bytes on noise. Mild penalty.
  \item \textbf{info-loss} --- SUMMARIZE when KEEP was best (verbatim values lost), OR DROP when SUMMARIZE was best (high-level info lost). Heavy penalty.
  \item \textbf{SEVERE!} --- DROP when KEEP was best. A signal turn vanishes from context. Critical penalty.
\end{itemize}

For each input turn, classify its (CHOSEN, BEST) into one of the six cells.

\partheading{PART 3 --- Structural sub-checks (always apply)}

\textbf{(i) RECENT-TURN PROTECTION.} Any turn within the last \texttt{max(5, R/3)} positions appearing in \texttt{output.summarize} or \texttt{output.drop} is a HARD VIOLATION $\rightarrow$ axis is capped at 1.

\textbf{(ii) CROSS-REFERENCE INTEGRITY} (Concern 4).
\begin{itemize}[leftmargin=*, itemsep=1pt]
  \item[(a)] For each entry in \texttt{output.unresolved} with field \texttt{\{turn: N\}}, verify $N$ is NOT in \texttt{output.drop}. A reference to a dropped turn is broken --- the cited turn no longer exists in the rebuilt context.
  \item[(b)] For each summary in \texttt{output.summarize}, verify its content is not duplicated by an entry in \texttt{output.unresolved}. The unresolved index is meant for OPEN issues only; if a summary already conveys the same info, the unresolved entry is redundant (or vice versa).
  \item[(c)] Multiple summaries / unresolved entries should not contradict each other on shared facts (e.g. two summaries giving different line numbers for the same function).
\end{itemize}
Each integrity violation subtracts.

\textbf{(iii) BUDGET COMPLIANCE} (Concern 5). Estimate the post-curation context size:

\begin{lstlisting}[language=jsonPrompt, basicstyle=\ttfamily\small]
post = sum(len(action) + len(observation) for kept turns)
     + sum(len(content) for summaries)
     + sum(len(summary) for unresolved entries)
     + len(curated_last_observation)
\end{lstlisting}

If \texttt{input.total\_tokens > input.budget} AND \texttt{post / CHARS\_PER\_TOKEN > input.budget}, the curator failed to bring the context under budget. Cap axis at 2. If \texttt{input.total\_tokens <= input.budget}, this sub-check is N/A.

\partheading{PART 4 --- Composite axis score}

Let \texttt{H\_eligible} = number of turns OUTSIDE the protected recent window. Tally:
\begin{itemize}[leftmargin=*, itemsep=1pt]
  \item \texttt{N\_correct}     = turns where CHOSEN == BEST
  \item \texttt{N\_severe}      = turns marked SEVERE! (DROP when KEEP was best)
  \item \texttt{N\_infoloss}    = turns marked info-loss
  \item \texttt{N\_undercomp+}  = turns marked under-comp+
  \item \texttt{N\_other}       = turns marked under-comp / over-cost
\end{itemize}

\begin{itemize}[leftmargin=*, itemsep=2pt]
  \item \textbf{5}: \texttt{N\_correct / H\_eligible >= 0.95} AND \texttt{N\_severe == 0} AND \texttt{N\_infoloss == 0} AND no integrity violations AND budget OK AND recent-turn clean.
  \item \textbf{4}: \texttt{0.85 <= ratio < 0.95} AND \texttt{N\_severe == 0} AND \texttt{N\_infoloss <= 1} AND no integrity violations AND budget OK.
  \item \textbf{3}: \texttt{0.70 <= ratio < 0.85} AND \texttt{N\_severe == 0}; OR exactly one info-loss; OR one minor integrity violation.
  \item \textbf{2}: \texttt{ratio < 0.70}; OR multiple info-loss; OR a cross-reference violation; OR budget compliance failed.
  \item \textbf{1}: \texttt{N\_severe >= 1} (any DROP-when-KEEP-best); OR recent-turn violation; OR most decisions are wrong.
\end{itemize}

\textbf{GUIDANCE for the rationale}: name 1--2 specific turns whose (CHOSEN, BEST) classification drove the score, including their turn numbers and the value or noise that justified the BEST action.

\axisheading{4. FIDELITY (1--5) --- serves Concern 2(b), part 1}

For every entry in \texttt{output.summarize}, compare the summary text against the corresponding raw turn in \texttt{input.history}:

\begin{itemize}[leftmargin=*, itemsep=2pt]
  \item \textbf{5}: Every value that appears in the raw turn AND matters for the task is preserved verbatim. No paraphrase of identifiers. The exact values to check: file paths (e.g. \texttt{django/db/models/fields/files.py}), line numbers (e.g. \texttt{276}), error codes (e.g. \texttt{E0401}, \texttt{ImportError}), function names (e.g. \texttt{deconstruct}), variable names, numeric outputs from tests/asserts, exception class names.
  \item \textbf{4}: One value paraphrased (``around line 50'' instead of ``line 47''), but no error codes or paths lost.
  \item \textbf{3}: Values mostly preserved; one minor identifier missing.
  \item \textbf{2}: An error code, traceback path, or function name was paraphrased or dropped.
  \item \textbf{1}: Summaries read like high-level descriptions with no verbatim values --- the agent could not act on them.
\end{itemize}

N/A $\rightarrow$ 5 if \texttt{output.summarize} is empty.

\axisheading{5. UNRESOLVED\_COVERAGE (1--5) --- serves Concern 1}

Two-direction check.

\textbf{(i) Existence.} For each entry in \texttt{output.unresolved}:
\begin{itemize}[leftmargin=*, itemsep=1pt]
  \item The cited turn really contains the claimed issue.
  \item The issue is GENUINELY OPEN at this turn --- not resolved by any later turn in \texttt{input.history}.
  \item The issue is a real ``issue'' (failing test, error message, unverified hypothesis, observation contradicting current approach) --- not a benign fact, completed action, or normal exploration step.
  \item Exact values are preserved (file paths, line numbers, error codes, function names).
\end{itemize}

\textbf{(ii) Coverage.} Scan \texttt{input.history} for open issues NOT in \texttt{output.unresolved}. Penalize for missing major open issues (failing test the agent has not addressed; error message not yet diagnosed; hypothesis not yet verified). Do not penalize for benign omissions.

\begin{itemize}[leftmargin=*, itemsep=2pt]
  \item \textbf{5}: Every entry is a real open issue with verbatim values; no major open issue is missing.
  \item \textbf{4}: One minor issue: one slightly paraphrased value, OR one non-critical open issue not listed.
  \item \textbf{3}: 2--3 minor issues; the index is still net useful.
  \item \textbf{2}: A listed unresolved was actually resolved earlier in the history, OR a major open issue is missing, OR an entry fabricates a value not in the cited turn.
  \item \textbf{1}: Most entries are fabricated, already-resolved, or missing their exact values, OR multiple major open issues are absent.
\end{itemize}

N/A $\rightarrow$ 5 ONLY when \texttt{output.unresolved} is empty AND the input history truly has no open issues. If \texttt{output.unresolved} is empty but the input has clear open issues, this axis scores 1--2 --- passthrough cannot mean ignoring open issues.

\axisheading{6. LAST\_OBS\_CURATION (1--5)}

Compare \texttt{curated\_last\_observation} to the input's \texttt{last\_observation}.

\begin{itemize}[leftmargin=*, itemsep=2pt]
  \item \textbf{5}: Short/clean input returned unchanged. Verbose input had banners / progress bars / duplicate lines trimmed but every error message, traceback, file path, line number, and numeric value preserved verbatim.
  \item \textbf{3}: Trimmed but lost one minor value, or unnecessarily trimmed clean input.
  \item \textbf{1}: Errors/values dropped, OR curated version is longer than the original.
\end{itemize}

\axisheading{7. INFORMATION\_DENSITY (1--5) --- serves Concern 2(b), part 2}

For each entry in \texttt{output.summarize}:

\begin{itemize}[leftmargin=*, itemsep=2pt]
  \item \textbf{5}: Every summary is 1--3 sentences and information-dense. Each states what was tried, what resulted, whether it advanced the task.
  \item \textbf{4}: One summary slightly off (4 sentences, or 15 chars) but the rest are fine.
  \item \textbf{3}: Summaries mostly fine, one is too long (\textgreater 4 sentences) or too short (\textless 10 chars / trivial).
  \item \textbf{2}: Multiple summaries fail length norms.
  \item \textbf{1}: Summaries are \textgreater 5 sentences each, OR most are \textless 20 chars and contentless.
\end{itemize}

N/A $\rightarrow$ 5 if \texttt{output.summarize} is empty.

\promptheading{Overall Score}

\begin{lstlisting}[language=jsonPrompt, basicstyle=\ttfamily\small]
overall = round( (worthiness*1.5 + schema*1 + decision*2.5 +
                 fidelity*2 + unresolved*1.5 + last_obs*1 +
                 density*1) / 10.5, 2 )
\end{lstlisting}

Decision is the highest weight (2.5) since it covers concerns 2(a), 3, and 4. Fidelity is 2 (verbatim values are critical to summary usefulness). Unresolved is 1.5 (Concern 1 in full). Schema, \texttt{last\_obs}, density are floors.

\promptheading{Output Format}

Emit only this JSON, nothing else:

\begin{lstlisting}[language=jsonPrompt]
{
  "input_worthiness": <int 1-5>,
  "schema": <int 1-5>,
  "decision_appropriateness": <int 1-5>,
  "fidelity": <int 1-5>,    // 5 if no summaries
  "unresolved_coverage": <int 1-5>,    // 5 only if no open issues
  "last_obs_curation": <int 1-5>,
  "information_density": <int 1-5>,    // 5 if no summaries
  "overall":  <float>,
  "keep_for_sft": <bool>,
  "rationale": "<2-4 sentences. Sentence 1: name 1-2 specific turn numbers whose (CHOSEN, BEST) classification drove the decision_appropriateness score - include the value or noise that justified BEST. Sentences 2-4: cite specifics for any other axis below 4>",
  "flags": ["<short tag>", ...]
    // possible flags:
    //   --- per-turn errors (Concern 1) ---
    //   "severe_drop" - DROP when KEEP was best (signal lost)
    //   "info_loss_summary"        - SUMMARIZE when KEEP was best
    //   "info_loss_drop"           - DROP when SUMMARIZE was best
    //   "missed_summarize"         - KEEP when SUMMARIZE was best (verbose turn left in)
    //   "missed_drop"              - KEEP when DROP was best (noise left in)
    //   "over_cost_summary"        - SUMMARIZE when DROP was best (wasted bytes)
    //   --- summary quality (Concern 3) ---
    //   "paraphrased_value"        - summary lost an exact value
    //   "trivial_summary"          - summary <20 chars / contentless
    //   "verbose_summary"          - summary >5 sentences
    //   --- unresolved (Concern 2) ---
    //   "fabricated_unresolved"    - unresolved cites non-existent issue
    //   "stale_unresolved"         - unresolved lists already-resolved issue
    //   "missed_open_issue"        - input has open issue not in unresolved
    //   --- cross-reference (Concern 4) ---
    //   "unresolved_cites_dropped" - unresolved entry points at a turn in drop[]
    //   "summary_unresolved_dup"   - summary duplicates an unresolved entry
    //   "contradictory_summaries"  - two summaries give conflicting values
    //   --- budget (Concern 5) ---
    //   "budget_overshot"          - post-curation size > budget on heavy input
    //   --- structural ---
    //   "dropped_recent_turn"      - touched a protected recent turn
    //   "trivial_passthrough"      - every turn KEEP on tiny input (low INPUT_WORTHINESS)
}
\end{lstlisting}

\end{promptbox}

\end{document}